\documentclass[12pt]{iopart}

\usepackage{iopams}  
\usepackage{threeparttable}
\usepackage{xcolor}
\usepackage{array}
\newcolumntype{u}[1]{>{\centering\arraybackslash}p{#1}}
\newcommand{\ineq}[1]{\footnotesize$#1$\normalsize}{}
\usepackage{graphicx}
\usepackage{subfig}
{}

\begin{document}

\title{Implementing Spiking Neural Networks on Neuromorphic Architectures: A Review}

\author{Phu Khanh Huynh, M. Lakshmi Varshika, Ankita Paul, Murat Isik, Adarsha Balaji, Anup Das}
%\author{Phu Khanh Huynh, M. Lakshmi Varshika, Ankita Paul, Murat Isik, Adarsha Balaji, Shihao Song, Anup Das}

\address{Drexel University, Philadelphia, PA 19104, USA}
\ead{anup.das@drexel.edu}
\vspace{10pt}
% \begin{indented}
% \item[]August 2017
% \end{indented}

\begin{abstract}
Recently, both industry and academia have proposed several different neuromorphic systems to execute machine learning applications that are designed using Spiking Neural Networks (SNNs). 
With the growing complexity on design and technology fronts, programming such systems to admit and execute a machine learning application is becoming increasingly challenging. 
Additionally, neuromorphic systems are required to guarantee real-time performance, consume lower energy, and provide tolerance to logic and memory failures.
Consequently, there is a clear need for system software frameworks that can implement machine learning applications on current and emerging neuromorphic systems, and simultaneously address performance, energy, and reliability.
Here, we provide a comprehensive overview of such frameworks proposed for 
%neuromorphic systems, addressing two key system design concepts -- 
both, platform-based design and hardware-software co-design.
%, and design-technology co-optimization.
We highlight challenges and opportunities that the future holds in the area of system software technology for neuromorphic computing.
\end{abstract}

%\maketitle

\section{Introduction}\label{sec:introduction}
Neuromorphic systems are integrated circuits designed to mimic the event-driven computations in a mammalian brain~\cite{mead1990neuromorphic}.
They enable execution of Spiking Neural Networks (SNNs), which are computation models designed using spiking neurons and bio-inspired learning algorithms~\cite{maass1997networks}.
SNNs enable powerful computations due to their spatio-temporal information encoding capabilities~\cite{paugam2012computing}.
SNNs can implement different machine learning approaches such as supervised learning~\cite{tavanaei2019deep}, unsupervised learning~\cite{dan2004spike}, reinforcement learning~\cite{lee2008synaptic}, and lifelong learning~\cite{schmidgall2021stable}.

\begin{figure}[h!]
	\centering
	%\vspace{-5pt}
	\centerline{\includegraphics[width=0.99\columnwidth]{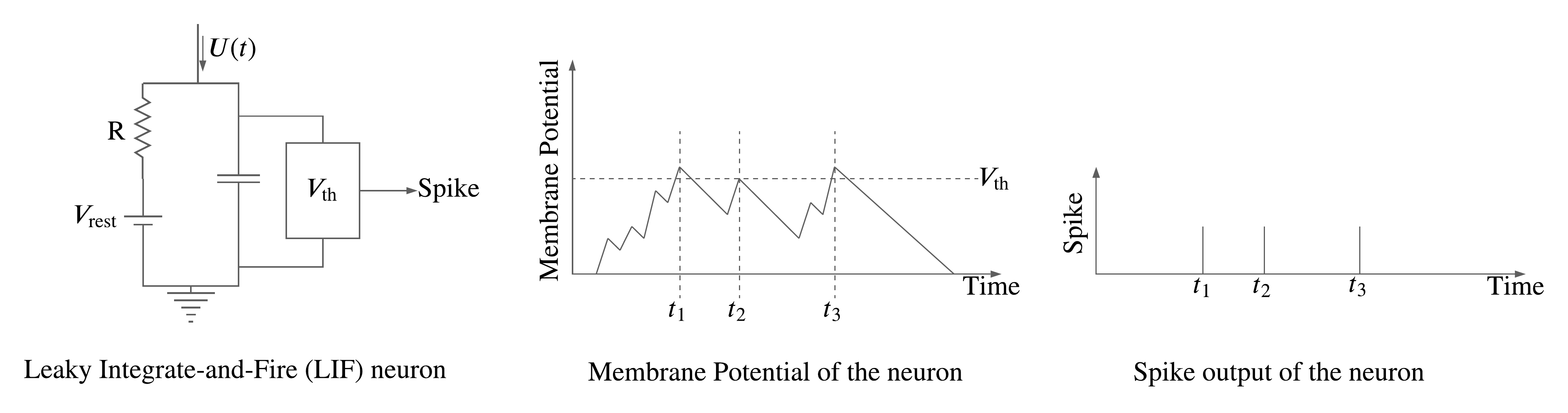}}
	%\vspace{-10pt}
	%\caption{An example of spiking neural network.}
	\caption{A leaky integrate-and-fire (LIF) neuron with current input $U(t)$ (left). The membrane potential over time of the neuron (middle). The spike output of the neuron representing its firing times (right).}
	%\vspace{-5pt}
	\label{fig:lif}
\end{figure}

In an SNN, neurons are connected via synapses. 
A neuron can be implemented as an integrate-and-fire (IF) logic~\cite{burkitt2006review}, which is illustrated in Figure~\ref{fig:lif} (left). 
Here, an input current spike \ineq{U(t)} from a pre-synaptic neuron raises the membrane voltage of a post-synaptic neuron. 
When this voltage crosses a threshold \ineq{V_{th}}, the IF logic emits a spike, which propagates to is post-synaptic neuron. 
Figure~\ref{fig:lif} (middle) illustrates the membrane voltage due to input spike trains. 
Moments of threshold crossing, i.e., the firing times are illustrated in Figure~\ref{fig:lif} (right). 
%These are the firing times of the output neuron.

SNNs can be implemented on a CPU or a GPU.
However, due to their limited memory bandwidth, performance of SNNs on such devices is usually slow and the power overhead is high.
In SNNs, neural computations and synaptic storage are tightly integrated. 
They present a highly-distributed computing paradigm which cannot be leveraged by CPU and GPU devices.
A neuromorphic hardware can eliminate the performance and energy bottlenecks of CPUs and GPUs, thanks to their low-power analog and digital neuron designs, distributed in-place neural computation and synaptic storage architecture, and the use of Non-Volatile Memory (NVM) for high-density synaptic storage~\cite{burr2017neuromorphic,chakraborty2020pathways,saxena2021neuromorphic,musisi2021viability,kim2018emerging,rose2021system,date2021neuromorphic}. 
Due to their low energy overhead, a neuromorphic hardware can implement machine learning tasks on energy-constrained embedded systems and edge devices of the Internet-of-Things (IoT)~\cite{li2018learning}.

A neuromorphic hardware is implemented as a tiled-based architecture~\cite{rajendran2019low}, where tiles are interconnected via a shared interconnect.
A tile may include 1) a neuromorphic core, which implements neuron and synapse circuitries, 2) peripheral logic to encode and decode spikes into Address Event Representation (AER), and 3) a network interface to send and receive AER packets from the interconnect. 
Switches are place on the interconnect to route AER packets to their destination tiles.
Table~\ref{tab:hw_examples} illustrates the capacity of some recent neuromorphic hardware cores.

\begin{table}[h!]
	\renewcommand{\arraystretch}{1.7}
	\setlength{\tabcolsep}{3pt}
	\caption{Capacity of some recent neuromorphic systems~\cite{christensen2021roadmap}.}
	\label{tab:hw_examples}
	%\vspace{10pt}
	\centering
	\begin{threeparttable}
	{\fontsize{8}{10}\selectfont
	    %\vspace{-10pt}
		\begin{tabular}{c|cccccccc}
			\hline
			& \textbf{ODIN} & $\mathbf{\mu}$\textbf{Brain} & \textbf{DYNAPs} & \textbf{BrainScaleS} & \textbf{SpiNNaker} & \textbf{Neurogrid} & \textbf{Loihi} & \textbf{TrueNorth}\\
			& \cite{odin} & \cite{mubrain} & \cite{dynapse} & \cite{brainscale} & \cite{spinnaker} & \cite{neurogrid} & \cite{loihi} & \cite{truenorth}\\
			\hline
			\textbf{\# Neurons/core} & 256 & 336 & 256 & 512 & 36K & 65K & 130K & 1M\\
			\textbf{\# Synapses/core} & 64K & 38K & 16K & 128K & 2.8M & 8M & 130M & 256M\\
			\textbf{\# Cores/chip} & 1 & 1 & 1 & 1 & 144 & 128 & 128 & 4096\\
			\hline
			\textbf{\# Chips/board} & 1 & 1 & 4 & 352 & 56 & 16 & 768 & 4096\\
			\hline
			%& \multicolumn{8}{|c}{High-performance neuromorphic system}\\
			\textbf{\# Neurons} & 256 & 336 & 1K & 4M & 2.5B & 1M & 100M & 4B\\
			\textbf{\# Synapses} & 256 & 336 & 65K & 1B & 200B & 16B & 100B & 1T\\
			\hline
	\end{tabular}}
	\end{threeparttable}
	%\vspace{12pt}
	%\vspace{-10pt}
\end{table}

NVM devices present an attractive option for implementing synaptic storage due to their demonstrated potential for low-power multilevel operations and high integration densities~\cite{jeong2018memristor,bez2004non,suzuki2015survey,nishi2011challenges}.
Recently, several NVMs are being explored for neuromorphic computing: Oxide-based Resistive Random Access Memory (ReRAM)~\cite{wong2012metal}, Phase Change Memory (PCM)~\cite{wong2010phase}, Ferroelectric RAM~\cite{arimoto2004current}, and Spin-Transfer Torque Magnetic or Spin-Orbit-Torque RAM (STT- and SoT-MRAM)~\cite{huai2008spin}. Table~\ref{tab:nvm_neuromorphic} shows some recent neuromorphic hardware demonstrations integrating NVMs.\footnote{Beside neuromorphic computing, NVMs are also used as main memory for conventional computing~\cite{lee2010phase,palp,datacon,mneme,hebe,kultursay2013evaluating,lee2009architecting,qureshi2009scalable}.}

\begin{table}[h!]
	\renewcommand{\arraystretch}{2.0}
	\setlength{\tabcolsep}{3pt}
	\caption{Neuromorphic hardware integrating NVMs.}
	\label{tab:nvm_neuromorphic}
	\vspace{10pt}
	\centering
	\begin{threeparttable}
	{\fontsize{8}{10}\selectfont
	    %\vspace{-10pt}
		\begin{tabular}{c|c}
			\hline
			\textbf{NVM Technology} & \textbf{References}\\
			\hline
			PCM & \cite{nandakumar2018phase} \\
			ReRAM & \cite{garbin2015hfo} \\
			FeRAM & \cite{jerry2017ferroelectric}\\
			STT-MRAM & \cite{vincent2015spin}\\
			%SOT-MRAM & \cite{}\\
			\hline
	\end{tabular}}
	\end{threeparttable}
	%\vspace{12pt}
	%\vspace{-10pt}
\end{table}

%phase-change memory (PCM)~\cite{wong2010phase}, spin-transfer torque magnetic memory (STT-MRAM)\cite{kultursay2013evaluating}, metal-oxide resistive memory (Re\-RAM)\cite{akinaga2010resistive}, conductive bridging memory (CB\-RAM)\cite{kund2005conductive}, ferro-electric memory (FeRAM)\cite{bondurant1990ferroelectronic}, etc.

\begin{figure}[h!]
	\centering
	\vspace{-5pt}
	\centerline{\includegraphics[width=0.99\columnwidth]{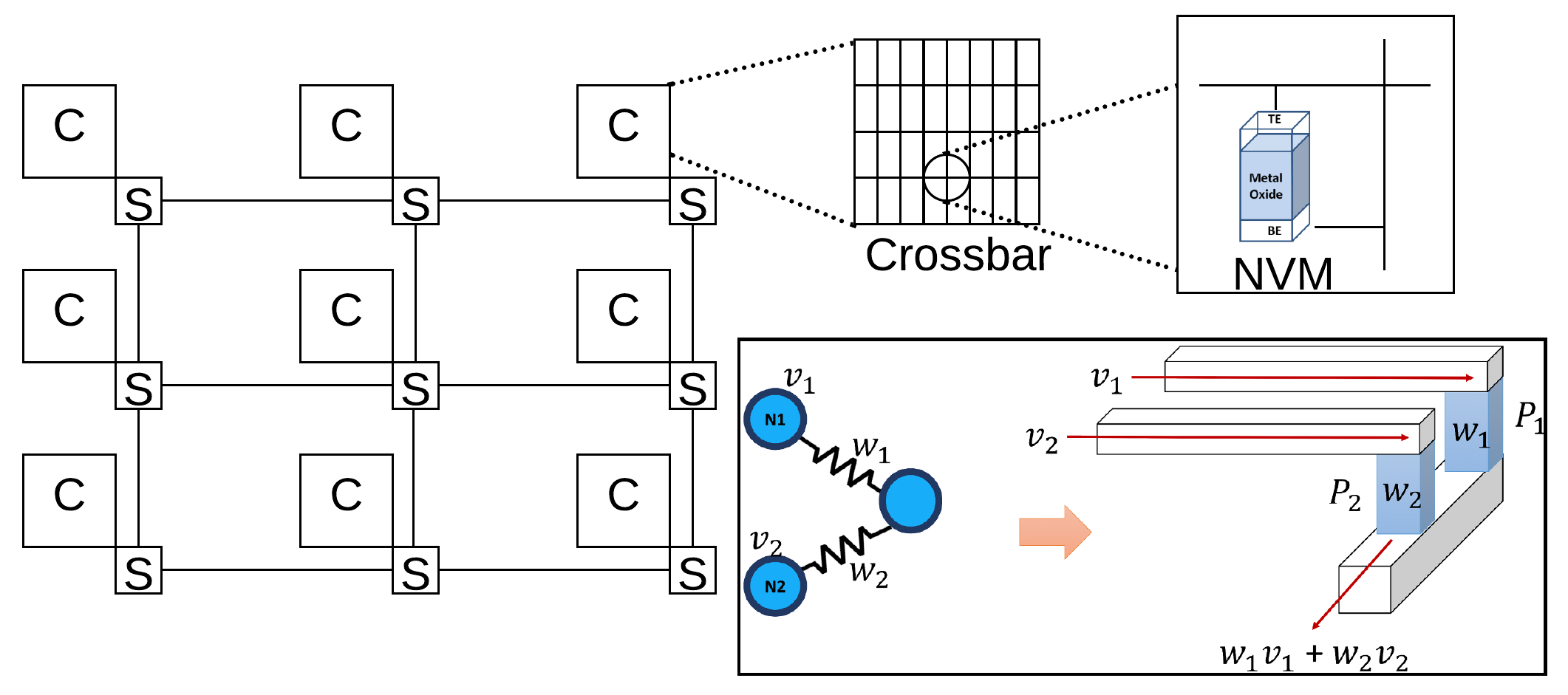}}
	\vspace{-5pt}
	\caption{A representative tile-based neuromorphic hardware~\cite{reneu}.}
	\vspace{-10pt}
	\label{fig:neuromorphic_hardware}
\end{figure}

Figure~\ref{fig:neuromorphic_hardware} shows a neuromorphic hardware with tiles (C) and switches (S). For illustration purposes, we show each tile as a crossbar, where NVM cells are organized in a two dimensional grid formed using horizontal wordlines and vertical bitlines.

%When implementing an SNN on a neuromorphic hardware, synaptic weights are programmed as conductance of NVM cells. 
The figure also illustrates a small example of implementing an SNN on a crossbar. 
Synaptic weights \ineq{w_{1}} and \ineq{w_{2}} are programmed as conductance of NVM cells P1 and  P2, respectively. The output spike voltages, \ineq{v_1} from N1 and \ineq{v_2} from N2, inject currents into the crossbar, which are obtained by multiplying a pre-synaptic neuron's output spike voltage with the NVM cell's conductance (Ohm's law). Current summations along columns are performed in parallel (Kirchhoff’s current law), and they implement the sum $\sum_j w_{i}v_i$ (i.e., neuron excitations).

To cope with the growing complexity of neuromorphic systems, challenges in integrating emerging NVM technologies, and faster time-to-market pressure, efficient design methodologies are needed. 
%We present three key concepts for neuromorphic system design, borrowed 
We highlight the following two key concepts
%from the Embedded Systems world 
that are likely to address the design issues postulated above.
\begin{itemize}
    \item \textbf{Platform-based Design:} In this design methodology, a hardware platform is abstracted from its system software, making the hardware and software developments orthogonal to allow a more effective exploration of alternative solutions~\cite{keutzer2000system}. 
    Platform-based design methodology facilitates the reuse of the system software for many different hardware platforms.
    \item \textbf{Hardware-Software Co-design:} In this design methodology, a hardware platform and its system software are concurrently designed to exploit their synergism in order to achieve system-level design objectives~\cite{de1997hardware}. The system software in this case is tailored for the hardware platform.
    %\item \textbf{Design-Technology Co-optimization:} In this design meth\-odology, system-level design metrics are applied to explore the choices in hardware design and process technology to enable scaling at advanced technology nodes~\cite{yeric2013past}.
\end{itemize}

In this paper, we provide a survey of these design methodologies for neuromorphic computing, focusing mainly on the recent advances on the software technology front.

%Executing SNN-based applications to a neuromorphic hardware is challenging.
%Since each core in a neuromorphic hardware can accommodate only a limited number of neurons and synapses, an SNN model needs to be first partitioned into clusters, where each cluster can be implemented on a core of the hardware. Partitioned clusters are then mapped to different cores using a run-time manager when admitting the model to the hardware platform. 

\section{Platform-based Design}\label{sec:pbd}
Platform-based design has emerged as an important design style for the electronics industry~\cite{keutzer2000system,sangiovanni2004benefits,sangiovanni2001platform,nuzzo2015platform,singh2013energy,das2018reliable}.
Platform-based design separates parts of the system design process such that they can be independently optimized for different metrics such as performance, power, cost, and reliability.
Platform-based design methodology can also be adopted for neuromorphic system design~\cite{neuroxplorer}, where the software can be optimized independently from the underlying neuromorphic hardware platform.

As in a conventional computing system, the abstractions for a neuromorphic system include 1) the \textit{application software}, 2) the \textit{system software}, and 3) the \textit{hardware}~\cite{jerraya2006programming,patterson2016computer,das2015hardware}.
In the context of neuromorphic computing, the application software includes applications designed using different SNN topologies such as mult-layer perceptron (MLP)~\cite{das2018heartbeat}, convolutional neural network (CNN)~\cite{moyer2020machine} and recurrent neural network (RNN)~\cite{HeartEstmNN}, and bio-inspired learning algorithms such spike timing dependant plasticity (STDP)~\cite{stdp}, long-term plasticity (LTP)~\cite{ltp}, and FORCE~\cite{force_snn}.
The system software includes the equivalent of a compiler and a run-time manager to execute SNN applications on the hardware. 
Finally, the hardware abstraction includes the platform, which consists of a neuromorphic hardware.

We focus on the system software abstraction and highlight key optimization techniques proposed in literature. 
In Section~\ref{sec:core_concepts}, we provide an overview of the uniqueness of system software concepts for neuromorphic computing.

For neuromorphic systems, performance, energy, and reliability are the driving metrics for the system software optimization.
Therefore, we categorize state-of-the-art approaches into 1) performance/energy-oriented software approaches (see Section~\ref{sec:perf_energy_pbd}) and 2) thermal/reliability-oriented software approaches (see Section~\ref{sec:thermal_reliability_pbd}).
Finally, we highlight recent approaches that use high-level dataflow representations to estimate SNN performance early in the platform design stage (see Section~\ref{sec:pbd_abstract}).

\subsection{System Software Considerations for Neuromorphic Computing}\label{sec:core_concepts}
In an SNN, information is encoded in spikes that are communicated between neurons~\cite{kiselev2016rate}.
We take the example of Inter Spike Interval (ISI) coding in SNNs.
Let  \ineq{\{t_1,t_2,\cdots,t_{K}\}} denote a neuron's firing times in the time interval \ineq{[0,T]}, the average ISI of this spike train is
\begin{equation}
    \label{eq:isi}
    \footnotesize \mathcal{I} = \sum_{i=2}^K (t_i - t_{i-1})/(K-1).
\end{equation}

\begin{figure}[h!]
	\centering
	%\vspace{-5pt}
	\centerline{\includegraphics[width=0.99\columnwidth]{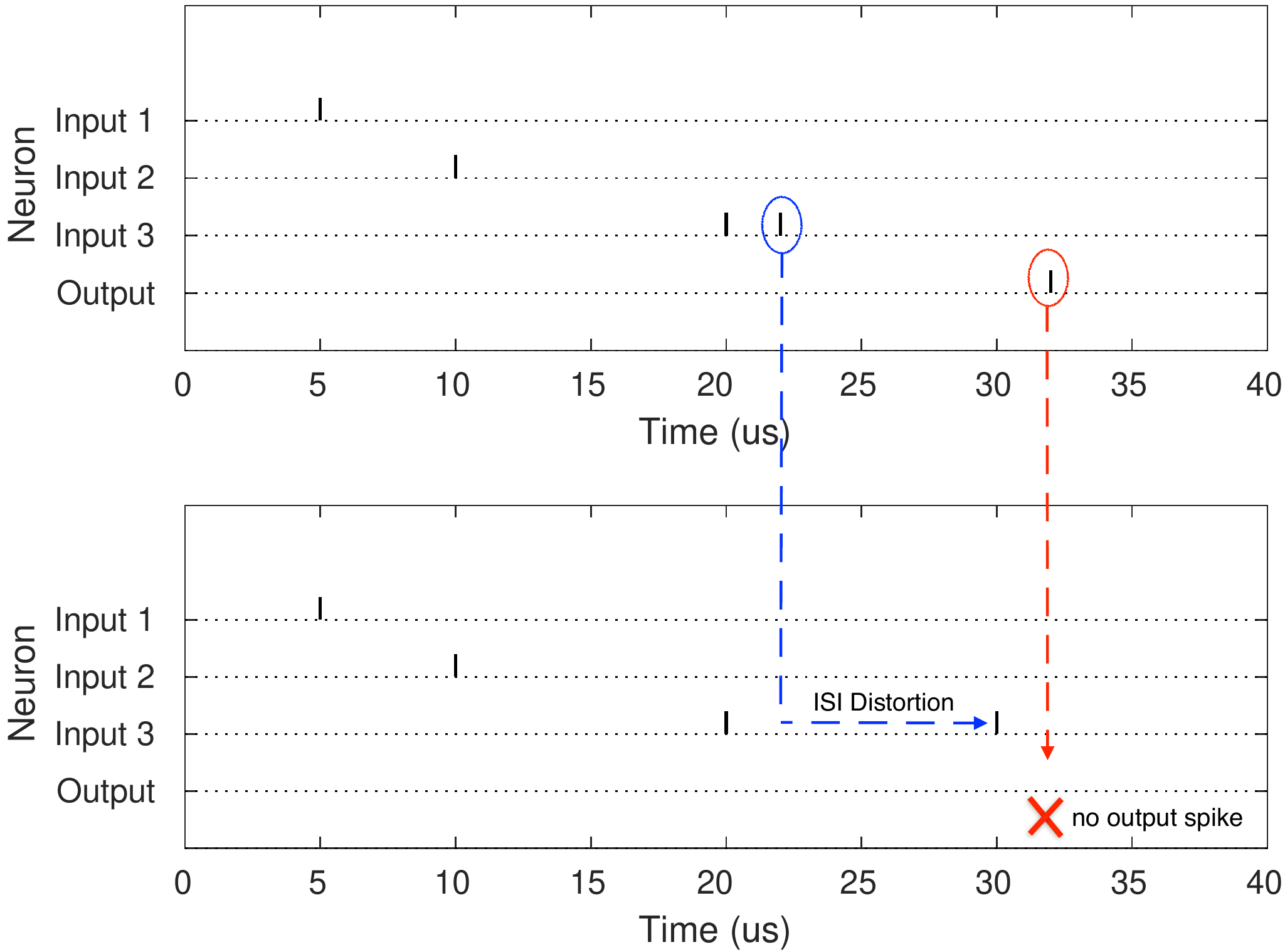}}
	%\vspace{-10pt}
	%\caption{An example of spiking neural network.}
	\caption{Impact of ISI distortion on accuracy. Top sub-figure shows a scenario where an output spike is generated based on the spikes received from the three input neurons. Bottom sub-figure shows a scenario where the second spike from neuron 3 is delayed. There are no output spikes  generated~\cite{pycarl}.}
	%\vspace{-5pt}
	\label{fig:isi_imact}
\end{figure}

A change in ISI, called \textit{ISI distortion}, impacts SNN performance.
To illustrate this, we use a small SNN in which three input neurons are connected to an output neuron. Figure~\ref{fig:isi_imact} illustrates the impact of ISI distortion on the output spike. In the top sub-figure, a spike is generated at the output neuron at 22$\mu$s due to spikes from input neurons. In the bottom sub-figure, the second spike from input 3 is delayed, i.e., it has an ISI distortion. Due to this distortion, there is no output spike generated. Missing spikes can impact inference quality, as spikes encode information in SNNs.

\begin{figure}[h!]%
    \centering
    %\vspace{-15pt}
    \subfloat[Original Image.\label{fig:original_image}]{{\includegraphics[width=4.0cm]{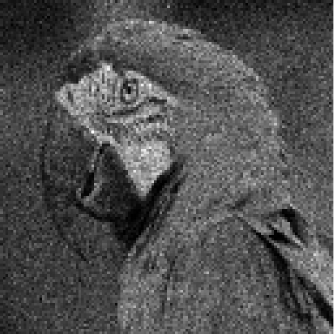} }}%
    %\quad
    %\hspace{0.001cm}
    \hfill
    \subfloat[Output with no ISI distortion (PSNR = 20).\label{fig:0ms_image}]{{\includegraphics[width=4.0cm]{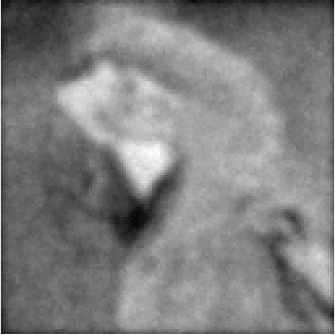}}}%
    \hfill
    \subfloat[Output with ISI distortion (PSNR = 19).\label{fig:20ms_image}]{{\includegraphics[width=4.0cm]{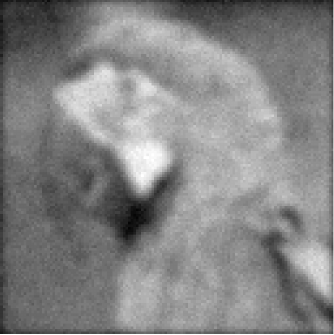}}}%
    %\vspace{-5pt}
    \caption{Impact of ISI distortion on image smoothing.}%
    \label{fig:ISI_impact}%
    %\vspace{-10pt}
\end{figure}

Figure~\ref{fig:ISI_impact} shows the impact of ISI distortion at the application level. We consider an image smoothing application implemented with an SNN model using the CARLsim simulator~\cite{carlsim}.
Figure~\ref{fig:original_image} shows the input image, which is fed to the SNN. Figure~\ref{fig:0ms_image} shows the output of the image smoothing application with no ISI distortion. Peak signal-to-noise ratio (PSNR) of the output with reference to the input is 20. Figure~\ref{fig:20ms_image} shows the output with ISI distortion. PSNR of this output is 19. A reduction in PSNR indicates that the output image quality with ISI distortion is lower. %In fact, image quality deteriorates with increase in ISI distortion. 

%\vspace{-10pt}

This background motivates the following. A system software for neuromorphic hardware needs to consider application property, especially the spike timing and their distortion to ensure that the SNN performance obtained on the hardware implementation matches closely to what is simulated in an application-level simulator such as CARLsim~\cite{carlsim}, NEST~\cite{nest}, Brian~\cite{brian}, and NEURON~\cite{neuron}.

\begin{figure}[h!]
	\centering
	%\vspace{-10pt}
	\centerline{\includegraphics[width=0.99\columnwidth]{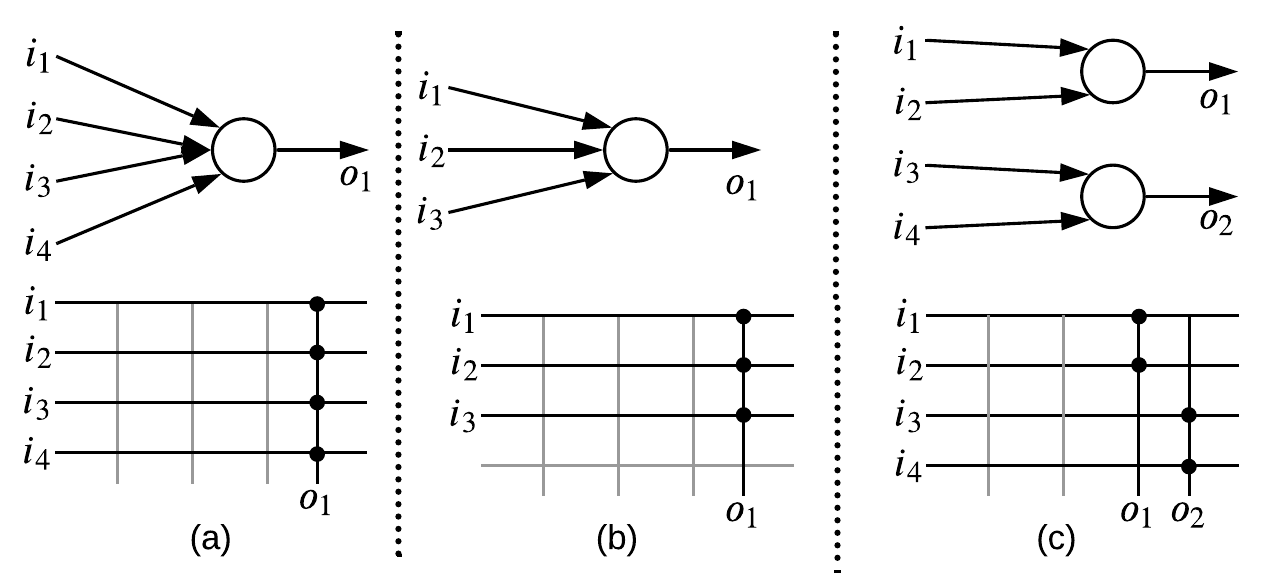}}
	%\vspace{-10pt}
	%\caption{An example of spiking neural network.}
	\caption{Implementation of a) one 4-input, b) one 3-input, and c) two 2-input neurons on a $4 \times 4$ crossbar~\cite{dfsynthesizer}.}
	\label{fig:crossbar_mapping}
\end{figure}

Implementation-wise, a system software for neuromorphic hardware must also incorporate hardware constraints, such as constrained neural architecture, limited neuron and synapse capacities, and limited fanin per neuron. 
We take the architecture of a crossbar~\cite{liu2015spiking,hu2014memristor,hu2016dot,ankit2017trannsformer,nukala2014spintronic,kim2012digital,zhang2018neuromorphic,gopalakrishnan2020hfnet,fernando20203d}, which is commonly used to implement neuromorphic hardware platforms.
In a crossbar, bitlines and wordlines are organized in a grid with memory cells connected at their crosspoints to store synaptic weights. 
Neuron circuitries are implemented along bitlines and wordlines.
A crossbar can accommodate only a limited number of pre-synaptic connections per post-synaptic neuron.
To illustrate this, Figure~\ref{fig:crossbar_mapping} shows three examples of implementing neurons on a $4 \times 4$ crossbar. The left sub-figure shows the implementation of a 4-input neuron on a crossbar. 
The neuron occupies all four input ports and one output port.
The middle sub-figure shows the implementation of a 3-input neuron, occupying three input ports and one output port. 
Finally, the right sub-figure shows the implementation of two two-input neurons, occupying four input ports and two output ports.

Crossbars are not the only type of neuromorphic architectures. 
The $\mu$Brain architecture consists of three layers of neurons interconnected in a fully-connected topology~\cite{mubrain}.
Both crossbars and $\mu$Brain suffer from limited fanin and fanout.
Other hardware architectures include decoupled design, where the processing logic is separated from the synaptic storage~\cite{pudiannao,dadiannao,chen2016eyeriss,chen2014diannao}.

When an SNN needs to be implemented in hardware, the system software needs to incorporate hardware constraints of the underlying platform.

\subsection{System Software for Performance and Energy Optimization}\label{sec:perf_energy_pbd}
We discuss key system software optimization approaches that address performance and energy aspects of executing SNN applications on a neuromorphic hardware.

In \cite{sentryos}, Varshika et al. propose a many-core hardware with fully-synthesizable clockless $\mu$Brain cores~\cite{mubrain}. 
In the proposed architecture cores are interconnected using a segmented bus interconnect~\cite{chen1999segmented}.
Internally, a $\mu$Brain core consists of three layers of fully-connected neurons that can be programmed to implement operations such as convolution, pooling, concatenation and addition, as well as irregular network topologies, which are commonly found in many emerging spiking deep convolutional neural network (SDCNN) models.
%The architecture of a $\mu$Brain core is different from a crossbar that is typically used in many neuromorphic hardware platforms. The design is shown to be scalable to several large-scale SDCNNs~\cite{cao2015spiking}.
To implement an SDCNN on the proposed many-core design, authors propose \underline{SentryOS}, a system software framework comprising of a compiler (SentryC) and a run-time manager (SentryRT).
SentryC is a clustering approach for SDCNNs to generate sub-networks that can be implemented on different cores of the hardware. 
The sub-network generation works as follows.
It sorts all neurons of an SDCNN model based on their distances from output neurons. 
It groups all neurons with distance less than or equal to 2 into clusters considering the resource constraint of a neuromorphic core. The constraint of 2 is due to the three-layered architecture of a $\mu$Brain core.
In the next iteration, it removes already clustered neurons from the model, recalculates neuron distances, and groups remaining neurons to generate the next set of clusters. 
The process is repeated until all neurons are clustered into sub-networks. 
By incorporating hardware constraints, SentryC ensures that a sub-network can fit onto the target core architecture.
The run-time manager (SentryRT) schedules these sub-networks onto cores. 
To do so, it uses a real-time calculus to compute the execution end times of different sub-networks.
Next, it discards execution times, retaining only the execution order of sub-networks.
Finally, it pipelines execution of sub-networks on cores and overlaps execution of multiple input images on to hardware cores.
By improving opportunities for pipelining and exploiting data-level parallelism, SentryOS is shown to significantly improve the hardware throughput.

In \cite{corelet}, Amir et al. propose \underline{corelet}, a programming paradigm for the TrueNorth neuromorphic hardware~\cite{truenorth}.
This is developed to address the complexity associated with designing neuromorphic algorithms that are consistent with the TrueNorth architecture and programming them on hardware.
The corelet paradigm is designed using \textit{Corelet}, an abstraction of a network of neurosynaptic cores that encapsulates biological details and neuron connectivity, exposing only a network's external inputs and outputs to the programmer.
Next, authors propose an object-oriented \textit{Corelet Language} for creating, composing, and decomposing corelets. 
It consists of three fundamental symbols -- neuron, neurosynaptic core, and corelet.
Connectors constitute the grammar for composing the symbols into TrueNorth program. 
Authors show that using symbols and grammar, it is possible to express any TrueNorth program.
Next, authors introduce the \textit{Corelet Library}, a repository of more than 100 corelets to facilitate designing TrueNorth programs. 
%The repository consists of more than 100 corelets.
Finally, authors propose a \textit{Corelet Laboratory} to implement programs designed using corelet onto the TrueNorth hardware.

%\cite{dey2021mapping}

In \cite{loihi_mapping}, Lin et al. propose \underline{LCompiler}, a compiler framework to map SNNs onto Loihi neuromorphic hardware~\cite{loihi}.
Authors show that the energy consumption in Loihi is mostly due to updating the data structures used by local learning rules.
Authors report that the energy consumed due to on-chip learning is an order of magnitude higher than the energy consumed in spike routing in Loihi.
Authors show that the energy cost associated in the learning updates is proportional to the number of data structures allocated for an SNN instance, which depends on how an SNN is partitioned into cores.
Internally, LCompiler creates a dataflow graph consisting of logical entities to describe an SNN instance. 
Such logical entities include compartments (main building block of neurons), synapses, input maps, output axons, synaptic traces, and dendritic accumulator.
Overall, LCompiler operates in three steps. 
In the pre-processing step, it validates SNN parameters, decomposes learning rules, transforms them into microcodes, and translates the SNN topology (i.e., logical entities) into a connection matrix.
In the mapping step, it uses a greedy algorithm to map logical entities to hardware components of Loihi cores, time-multiplexing the shared resources on a core.
Finally, in the code generation step, it generates the binary bitstream for each Loihi core.

In \cite{pacman}, Galluppi et al. propose the PArtitioning and Configuration MANager (\underline{PACMAN}), a framework to map SNNs to computational nodes on the SpiNNaker system~\cite{spinnaker}.
The key idea of PACMAN is to transform the high-level representation of an SNN to a physical on-chip implementation.
PACMAN holds three different representations of an SNN.
In the Model-Level, an SNN is specified using a high-level language such as PyNN~\cite{pynn} and Nengo~\cite{nengo}.
In the System-Level, an SNN is split into groups, where each group can fit onto a SpiNNaker computational node.
In the Device-Level, a mapping is formed from groups to nodes of the SpiNNaker system.
To generate these representations, PACMAN performs four operations -- 1) splitting, which partitions an SNN into smaller sub-networks, 2) grouping, which combines sub-networks to groups that can fit onto a node of the hardware, 3) mapping, which allocates groups to different nodes, and 4) binary file generation, which creates the actual data binary from a partitioned and mapped network.

In \cite{sugiarto2017optimized}, Sugiarto et al. propose a framework to map general purpose applications running as a task graph on to the SpiNNaker hardware, with the objective of reducing the data traffic between different SpiNNaker chips.
The proposed framework uses the task graph description given by XL-Stage program~\cite{campos2016xl}.
In mapping a task graph to the hardware, each task is mapped to a SpiNNaker chip.
To provide fault tolerance, multiple copies of a task are generated and executed on different SpiNNaker chips.
Authors propose to use an evolutionary algorithm to perform the mapping with the objective of balancing the load on the system, while minimizing inter-chip data communication.

In \cite{pynn}, Davidson et al. propose a Python interface called \underline{PyNN} to facilitate faster application development and portability across different research institutes.
PyNN provides a high-level abstraction of SNN models, promotes code sharing and reuse, and provides a foundation for simulator-agnostic analysis, visualization and data-management tools. 
Apart from serving as the Python front-end for different backend SNN simulators, PyNN supports mapping SNN models on the SpiNNaker~\cite{spinnaker}, BrainScaleS~\cite{brainscale}, and Loihi~\cite{loihi} neuromorphic hardware.
To do this mapping, PyNN first partitions an SNN model into clusters and then executes them on different cores of the hardware.
To do the partitioning, PyNN arbitrarily distributes neurons and synapses of an SNN model while incorporating the resource constraint of a neuromorphic core.

In \cite{nengo}, Bekolay et al. propose \underline{Nengo}, a Python framework for building, testing and deploying SNNs to neuromorphic hardware.
Nengo is based on the Neural Engineering Framework (NEF)~\cite{eliasmith2003neural}.
NEF uses three basic principles to construct large-scale neural models. 
These principles are 1) representation, 2) transformation, and 3) dynamics.
The representation principle of NEF proposes that information is encoded by population of neurons.
It represents information as time-varying vectors of real numbers.
To encode information, current is injected into neurons based on the vector being encoded.
The original encoded vector can be estimated using a decoding process, which involves filtering the spike trains using an exponentially decaying filter.
Filtered spike trains are summed together with weights that are determined by solving a least-square minimization problem.
The transformation principle of NEF proposes that the weight matrix between two neural populations can be factored into two significantly smaller matrices that are computationally efficient.
Finally, the dynamics principle states that with recurrent connections, vectors corresponding to neural populations are equivalent to state variables in a dynamic system.
Such a system can be analyzed using the control theory and translated into neural circuitry using the principles of representation and transformation.
Nengo uses NEF to not only simulate a large-scale neural model but also to map SNNs to custom FPGA (nengoFPGA) and on neuromorphic hardware such as Loihi~\cite{loihi} and SpiNNaker~\cite{spinnaker}.

In \cite{ji2018bridge}, Ji et al. propose a compiler to transform a trained SNN application to an equivalent network that satisfies the hardware constraints.
This compiler is targeted for the TianJi~\cite{tianji} and PRIME~\cite{prime} neuromorphic hardware platforms.
There are four steps to perform the compilation.
In step 1, the compiler builds a dataflow graph based on the given SNN information which includes trained parameters, network topology, vertex information, and training dataset.
In step 2, it transforms the dataflow graph into an intermediate representation (IR) which consists of hardware-friendly operations.
In step 3, it performs graph tuning, which includes data re-encoding (to address the hardware precision problem), expanding (where the IR is converted to operations supported in the target hardware), and weight tuning (i.e., fine tuning parameters to minimize transformation errors).
Finally in step 4, it maps the tuned graph to hardware, exploiting a platform's interconnection constraints.
The mapping step assigns hardware operations to physical cores of a hardware.

In \cite{gao2012dynamical}, Gao et al. propose an approach to map neuronal models to a neuromorphic hardware by exploiting dynamical system theory.
Authors take a model-guided approach to mapping neuronal models onto neuromorphic hardware.
The overall approach is as follows. 
It starts from the ordinary differential equations (ODEs) that govern a model's state variables.
Next it represents these state variables as currents.
Next, the current-mode subthreshold CMOS circuits are synthesized directly from these ODEs.
This approach yields combinations of circuit biases that are related to neural model parameters through a set of hardware-specific mapping parameters.
Finally, these mapping parameters are translated into circuit parameters.
Mapping parameters are extracted using 1) a non-linear optimization using the model's numerical simulation as reference, and 2) an iterative approach to adjust the circuit's operating current to converge with the numerical simulation's state variables.
Authors demonstrate their mapping strategy using quadratic and cubic Integrate and Fire neurons into the many-core Neurogrid neuromorphic hardware~\cite{neurogrid}.

In \cite{neftci2013synthesizing}, Neftci et al. propose an approach to mapping SNNs on neuromorphic hardware with imprecise and noisy neurons. 
To address this, authors propose to first transform an unreliable hardware layer of silicon neurons into an abstract computational layer composed of reliable neurons, and then modeling the target dynamics as a soft state machine running on this computational layer.
The mapping idea is the following.
An SNN is realized on hardware by mapping a neuron's circuit bias voltages to the model parameters and  calibrating them using a series of population activity measurements. 
The abstract computational layer is formed by configuring neural networks as generic soft winner-take-all sub-networks.
Finally, states and transitions of the desired high-level behavior are embedded in the computational layer by introducing only sparse connections between some neurons of the various sub-networks.

In \cite{psopart}, Das et al. propose Particle Swarm Optimization (PSO)-based SNN PARTitioning (\underline{PSOPART}), a technique to partition an SNN model into short-distance local synapses and long-distance global synapses for efficient mapping of the SNN to the many-core DYNAPs neuromorphic hardware~\cite{dynapse}.
Local synapses are mapped within different neuromorphic cores, while the global synapses are mapped on the shared interconnect.
The mapping problem is solved using PSO.
First, an optimization problem is formulated.
A set of neurons and a set of cores are assumed. A set of binary variables \ineq{x_{i,j}\in\{0,1\}} is  defined, where \ineq{x_{i,j}} assumes `1' if neuron \ineq{i} is mapped to core \ineq{j}.
Constraint to the problem is that each neuron can be mapped to only one core. 
Total number of variables generated is \ineq{D = N\times C}, where \ineq{N} is the number of neurons and \ineq{C} is the number of cores.
Objective of the optimization problem is to minimize the number of spikes communicated between different cores.
Next, the optimization problem is transformed into PSO domain.
To this end, PSOPART instantiates a population of swarms.
The position of each swarm is defined in a \ineq{D}-dimensional space.
PSOPART finds a solution to the optimization problem by iteratively moving these swarms in the search space while recording each swarm's local best position in relation to the global best solution~\cite{pso}.

%The partitioning is performed using an instance of the Particle Swarm Optimization (PSO) to minimize the number of spikes on global synapses. PSO is a meta-heuristic approach to find a solution to an optimization problem by iteratively trying to improve a candidate solution with regard to a given measure of quality~\cite{pso}.  

In \cite{spinemap}, Balaji et al. propose SPIking Neural NEtwork MAPping (\underline{SpiNeMap}), a framework that first partitions an SNN model into clusters and then places these clusters on different physical cores of a neuromorphic hardware.
SpiNeMap uses a greedy approach, roughly based on the Kernighan-Lin Graph Partitioning algorithm~\cite{kernighan1970efficient} to partition an SNN into clusters, minimizing the inter-cluster spike communication.
It then uses an instance of the PSO to place each cluster on a physical core of the neuromorphic hardware, where cores are arranged in a two-dimensional mesh.
The objective of the PSO is to minimize the number of hops on the shared interconnect that spikes need to communicate before they reach their destination.
In this way, the energy consumption on the shared interconnect is reduced.

In \cite{urgese2016optimizing}, Urgese et al. propose \underline{SNN-PP}, a partitioning and placement methodology to map SNNs on SpiNNaker parallel neuromorphic platform.
The objective of SNN-PP is to improve on-chip and off-chip communication efficiency.
The methodology is developed in two phases.
In the first phase, SNN-PP profiles the hardware architecture of SpiNNaker to detect bottlenecks in the communication system.
Next, SNN-PP describes an SNN application as a graph where each node represents a population, which is a homogeneous group of neurons sharing the same model and parameters.
Each edge is called projection, which connects neurons of two different populations using synaptic connections.
The graph partitioning step of SNN-PP minimizes communication between two populations.
Finally, populations are mapped to SpiNNaker cores using the Sammon Mapping algorithm~\cite{sammon1969nonlinear} to reduce inter-core spike communication.

In \cite{barchi2018directed,barchi2018mapping}, Barchi et al. propose a design methodology to map SNNs on globally asynchronous and locally synchronous (GALS) multi-core neuromorphic hardware such as SpiNNaker~\cite{spinnaker}.
Authors propose a task placement pipeline to minimize the spike communication between different computing cores.
The target is to reduce the distance spikes communicate before reaching their destination.
The methodology represents both an SNN and the platform as separate directed graphs.
On the application side, nodes are neurons or population of neurons and edges are synaptic connections.
On the platform side, nodes are processing cores while edges represent physical communication channels between them.
The mapping problem is formulated as the allocation of application nodes to the platform nodes with the objective of minimizing the inter-core spike communication.
To this end, authors propose to use simulated annealing (SA)~~\cite{van1987simulated} to explore different neuron-to-core mappings and evaluate the cost function.

%\cite{barchi2018mapping}

% In \cite{li2020sneap}, Li et al. propose a toolchain to map SNNs on neuromorphic hardware.
% This framework is based on the SpiNeMap approach~\cite{spinemap}.
% Similar to the SpiNeMap approach, the toolchain first partitions an SNN into groups, where a group of neurons and synapses can be implemented on a crossbar-based neuromorphic hardware.
% However, unlike SpiNeMap which uses an instance of PSO for mapping groups to hardware crossbars, authors propose to use TABU search and show the speedup obtained in mapping of SNNs to hardware.

In \cite{twisha_energy}, Titirsha et al. propose an energy-aware mapping of SNN models to crossbar-based neuromorphic hardware. 
The proposed approach first models the total energy consumption of a neuromorphic hardware, considering both static and dynamic powers consumed by neurons and synapses of an SNN model on hardware, and the energy consumed in communicating spikes on the shared interconnect.
This is an extension of the NCPower framework~\cite{wang2020ncpower}, which models only the dynamic energy consumption of a neuromorphic hardware.
Next, authors show that different SNN mapping strategies lead to a difference in the energy consumption on the hardware.
Finally, authors propose an iterative approach of mapping an SNN model to the hardware with the objective of minimizing the total energy consumption.
The iterative approach first partitions an SNN model into clusters, and then these clusters are randomly placed on cores of a hardware. 
Thereafter, clusters are repeatedly swapped on cores to see if there is any reduction in the energy consumption.
The iterative approach terminates when there is no energy reduction obtained during swapping.

In \cite{pycarl}, Balaji et al. propose \underline{PyCARL}, a Python frontend for the CARLsim-based SNN simulator~\cite{carlsim}.
CARLsim {facilitates} parallel simulation of large SNNs using CPUs and multi-GPUs, simulates multiple compartment models, 9-parameter Izhikevich and leaky integrate-and-fire (LIF) spiking neuron models, and integrates the fourth order Runge Kutta (RK4) method for improved numerical precision. CARLsim's support for built-in biologically realistic neuron, synapse, current and emerging learning models and continuous
integration and testing, make it an easy-to-use and powerful simulator of biologically-plausible SNN models. Benchmarking results demonstrate simulation of 8.6 million neurons and 0.48 billion synapses using 4 GPUs and
up to 60x speedup with multi-GPU implementations over a single-threaded CPU implementation.
PyCARL also integrates a cycle-accurate neuromorphic hardware simulator, which facilitates simulating a machine learning model on crossbar-based neuromorphic hardware platforms such as DYNAPs~\cite{dynapse} and TrueNorth~\cite{truenorth}.
Internally, PyCARL uses the SpiNeMap framework~\cite{spinemap} to map applications to the hardware simulator.
PyCARL allows users to estimate the accuracy deviation from software simulation that may result due to hardware latency.

%In \cite{eliasmith2003neural}, Eliasmith et al. propose \underline{NEF}, a Neural Engineering Framework to implement SNNs on custom neuromorphic hardware.

In \cite{corradi2014mapping}, Corradi et al. propose to use the {NEF} framework~\cite{eliasmith2003neural} to perform arbitrary mathematical computations using a mixed-signal analog/digital neuromorphic hardware.
Authors propose to translate these computations into the three 
%The NEF is based on control theory and it integrates three 
basic principles of NEF -- 1) Representation, 2) Transformation, and 3) Dynamics.
Through representation, authors propose to encode a stimulus as spiking activity of a group of neurons.
Through transformation, authors propose to convert the stimulus using a weighted connection between pools of neurons.% is used to compute a function on the represented value.
Through dynamics, authors propose to implement linear and non-linear dynamical models including attractor networks, Kalman filters, and controllable harmonic oscillators.

In \cite{decompose_snn}, Balaji et al. propose \underline{DecomposeSNN}, an approach to decompose each neuron of an SNN with many pre-synaptic connections to a sequence of homogeneous neural units (called Fanin-of-Two or FIT units), where each FIT unit can have a maximum of two pre-synaptic connections.
This spatial decomposition technique is proposed for crossbar-based neuromorphic hardware, where each crossbar can only accommodate a limited number of pre-synaptic connections, i.e., fanins per post-synaptic neuron.
The mapping framework works as follows. 
An SNN model is transformed into a representation consisting of FIT units.
Next, FIT units are packed into clusters incorporating a crossbar's fanin constraint.
In this way, a cluster can be incorporated directly on a crossbar.
Finally, clusters are mapped to crossbars of a neuromorphic hardware using the SpiNeMap framework~\cite{spinemap}.
%DecomposeSNN presents several advantages compared to conventional crossbar mapping approaches.
%irst, it allows mapping of FIT units from different post-synaptic neurons on to the same crossbar, improving crossbar utilization.
%Second, by decomposing a complex neuron into smaller FIT units, DecomposeSNN can map all pre-synaptic connections of every neuron in an SNN. Since it does not drop synaptic connections during crossbar mapping, DecomposeSNN improves SNN performance on hardware.

%\textcolor{red}{
In \cite{coreset}, Yang et al. propose the CoreSet Method (\underline{CSM})  to mitigate the limited fanin/fanout constraints of a neuromorphic crossbar in mapping SNNs to hardware.
The idea is based on coreset, which consists of multiple cores of particular arrangements.
CSM allows flexible aggregation of cores, allowing the chip to support different network models with improved resource efficiency.
CSM also supports different core sizes for optimal silicon area usage.
CSM is an end-to-end full-stack framework that includes TensorFlow-based training, coreset-based compilation, and FPGA emulation.
The compilation step splits and merges coresets to optimize the silicon area.
%}

In \cite{balaji2020run}, Balaji et al. propose a run-time manager for neuromorphic hardware.
The key motivation is that many machine learning models enable continuous learning where synaptic connections are updated during model execution.
Compile time-based mapping approaches such as SpiNeMap~\cite{spinemap} is not efficient for such dynamic scenarios.
To address this, authors propose a fast approach to adjust the mapping of a model to the hardware at run-time once a synaptic update is made to the model.
The proposed approach is based on a Hill-Climbing heuristic~\cite{selman2006hill}, which quickly finds a new mapping of neurons and synapses to different cores of the hardware with the objective of minimizing the inter-core spike communication.
The algorithm is an iterative approach, which starts with a random allocation of neurons and synapses to the hardware.
Subsequently, the algorithm iterates to find a better local solution by making an incremental change in the mapping such as relocating a neuron to a different core.
The objective is to reduce inter-core spike communication. 

In \cite{plank2018tennlab}, Plank et al. propose \underline{TENNLab}, an exploratory framework that provides the interface and software support for the development and testing of neuromorphic applications and devices.
The programming approach utilizes the Evolutionary Optimization of Neuromorphic Systems (EONS) for application development.
The software architecture of the framework includes three libraries and two driver programs.
The first library is the engine, which implements application-specific functions.
This is independent of neuromorphic devices.
The second library is the driver, which manages the interaction between the framework and neuromorphic devices.
The third library is the EONS library, it provides an interface between the graph representation of an application and the network structure supported by the hardware. 
There are two drivers -- a standalone driver for executing the application and the EONS driver to train a network for a given application and a neuromorphic device.
Users can edit these libraries and drivers of TENNLab to implement SNNs on a neuromorphic device.

In \cite{mitchell2021low}, Mitchell et al. propose an approach to improve the energy efficiency of SNN training using neuromorphic cores implemented on Xilinx Zynq platform.
To do so, authors develop a system software using the TENNLab framework~\cite{plank2018tennlab} targeted for the Caspian neuromorphic platform~\cite{mitchell2020caspian}.
%and it integrates with the TENNLab framework~\cite{plank2018tennlab}.
Using this software framework, authors accelerate and improve the energy efficiency of the evaluation step of an evolutionary algorithm used to train SNNs.
The framework works as follows.
First, the dataset is encoded into a sequence of spiking packets.
For each epoch, the interface receives a population of networks.
These networks are processed using a pipeline of thread pools. 
The first thread pool maps a network into configuration command.
The second pool sends configuration packets and dataset.
Finally, the third pool decodes the output.

%In \cite{tang2020real}, Tang et al. propose the mapping of simultaneous localization and mapping (SLAM) system in mobile robots. The proposed SNN structure consists of the Winner-Take-ALL (WTA) structure and heterosynaptic competitive learning for place field generation and dendritic trees for reference frame transformation. This SNN is implemented using Loihi neuromorphic hardware and interfaced using the ROS running on the mobile robot.

In \cite{balaji2019design}, Balaji et al. propose a design methodology called \underline{ROXANN} for implementing artificial neural networks (ANNs) using processing elements (PEs) designed with low-precision fixed-point numbers and high performance and reduced-area approximate multipliers in FPGAs.
The design methodology is the following.
First, authors propose the design of a generic processing element that supports the use of accurate/approximate arithmetic units, such as adders and multipliers, to compute the partial weighted sum of neurons~\cite{das2013improving,ullah2018smapproxlib,prabakaran2018demas,ullah2018area}.
Next, authors use quantized activation and weights during training and implementation phases of a neural network. 
Finally, authors propose a clustering-based approach, used after the training phase of the neural network, to minimize the number of distinct weights on the network.
Although ROXANN is developed for ANNs, authors discuss the modifications needed to apply ROXANN for SNNs.

In \cite{ankit2017trannsformer}, Ankit et al. propose \underline{TraNNsformer}, an integrated framework for training and mapping SNNs to crossbar-based neuromorphic hardware.
TraNNsformer works on the fully-connected layer as follows. 
First, it performs size-constrained iterative clustering to generate clusters from a connectivity matrix.
The algorithm is based on Spectral Clustering~\cite{ng2002spectral}, a graph clustering algorithm that produces a set of disjoint graph nodes such that intra-cluster associativity is maximized.
The proposed iterative spectral clustering minimizes the number of unclustered synapses while ensuring cluster generation with high utilization.
Finally, TraNNsformer retrains the network to fine-tune connections, while reinforcing the clustering.

In \cite{neunoc}, Liu et al. propose \underline{Neu-NoC}, an efficient interconnect to reduce the redundant data traffic in a neuromorphic hardware.
Authors first perform comprehensive evaluation of traditional Network-on-Chip (NoC) for neuromorphic system to identify design bottlenecks.
Next, authors propose a hierarchical NoC for neuromorphic hardware.
It consists of local rings for the layers of a machine learning model and a global mesh to interconnect the local links.
Next, authors propose an efficient mapping of machine learning models to the Neu-NoC-based neuromorphic hardware.
The mapping consists of placing the layers to local rings such that the distance of data communication is reduced, which reduces data congestion.
Finally, a multicast transmission is proposed to reduce the amount of data communication and a new type of flit is introduced to further reduce the data congestion.

In \cite{balaji2019exploration}, Balaji et al. propose a scalable neuromorphic design based on the DYNAPs neuromorphic hardware~\cite{dynapse} and using a Segmented Bus-based interconnect for spike communication.
Authors also propose mapping SNNs onto the target architecture.
Authors argue that mesh-based NoCs that are used to interconnect cores in recent neuromorphic designs,
%However, NoC interconnect has 
have relatively long time-multiplexed connections that need to be near-continuously powered up and down, reaching from the ports of data producers/consumers (inside a core or between different cores) up to the ports of communication switches~\cite{wang2012minimizing,yang2020recent,wasif2020energy,neunoc,yoon2017system,das2012fault}.
To address this, authors propose segmented bus. Here, a bus lane is partitioned into segments, where interconnections between segments are bridged and controlled by switches~\cite{chen1999segmented}. 
Authors propose a dynamic segmented bus architecture with multiple segmented bus lanes.
An optimized controller is designed to perform mapping of communication primitives to segments by profiling the communication pattern between different cores for a given SNN.
Based on this profiling and mapping, switches in the interconnect are programmed once at design-time before admitting an application to the hardware. 
By avoiding run-time routing decisions, the proposed design significantly gains on energy and latency.

In \cite{catthoor2018very}, Catthoor et al. propose a scalable neuromorphic hardware design to map large-scale SNN applications.
The key idea is to design a segmented bus-based interconnect with three way switches built using thin-film transistors (TFTs).
Authors propose to integrate these TFT switches into the back-end-of-line (BEOL) fabric.
%Authors show the reduction of energy through intelligent mapping of SNNs to the proposed hardware.
Authors propose to use the SpiNeMap framework~\cite{spinemap} to map SNN applications to the target architecture.

In \cite{jolpe18}, Balaji et al. propose a design methodology to perform heartbeat classification on an event-driven neuromorphic hardware such as the DYNAPs~\cite{dynapse}.
The methodology starts with an optimized CNN implementation of the heartbeat classification task.
It then converts CNN operations, such as multiply-accumulate, pooling and softmax, into spiking equivalent with a minimal loss of accuracy.
Finally, it performs power and accuracy tradeoffs by controlling the synaptic activity in the hidden layers using normalization. 
Authors implement the converted CNN on DYNAPs~\cite{dynapse} using the SpiNeMap framewprok~\cite{spinemap}.
%Compared to a standalone CNN implementation, the proposed methodology shows 87.7\% and 96.8\% reduction in energy consumption with only 0.6\% and 1.0\% lower accuracy. respectively.

In \cite{wang2021end}, Wang et al. propose an end-to-end framework for mapping hybrid neural networks involving ANN and SNN to the Tianjic neuromorphic hardware~\cite{pei2019towards}.
While intensive data representation of ANNs makes them achieve higher accuracy, event-driven spike trains of SNNs make them energy efficient.
Hybrid neural networks (HNNs) allow to combine the best of both worlds.
The proposed framework is used to implement the ANN module, SNN module, and 
%In the proposed end-to-end framework, the implementation of an HNN into ANN module, SNN module, and 
the signal conversion module between ANN and SNN.
The ANN computation module is mapped using~\cite{esser2013cognitive}.
The SNN computation module is mapped using~\cite{ji2018bridge}.
To implement the communication module, four different signal conversion methods are evaluated.
A global timing adjustment mechanism is also developed between these different modules.

In \cite{nair2019mapping}, Nair et al. propose mapping of recurrent neural networks (RNNs) to in-memory neuromorphic chips. 
%Authors propose an adaptive spiking neuron model that can be abstracted as a low-pass filter.
The mapping procedure is the following.
First, the recurrent ANN cell is modified by replacing the RNN units with an adaptive spiking neuron model that can be abstracted as a low-pass filter.
Second, the modified network is trained using Backprop~\cite{lecun2012efficient}.
This generates trained synaptic weights for the recurrent SNN.
Third, to ensure the spiking neurons do not saturate, the largest value attained by state variables in the trained network is mapped to the input.
Finally, inputs of the SNN are re-scaled to suitable current or voltage levels.

In \cite{shihao_soda}, Curzel et al. propose an automated framework called \underline{SODASNN} to synthesize a hybrid neuromorphic architecture consisting of digital and analog components.
The framework consists of the software defined architecture ({SODA}) synthesizer~\cite{minutoli2020soda}, a novel no-human-in-the-loop hardware generator that automates the creation of machine learning (ML) accelerators from high-level ML language.
Inside the SODA framework, authors implemented the machine learning intermediate representation (MLIR) dialect~\cite{lattner2020mlir}, which allows mapping spiking neurons and their computations to corresponding specialized hardware. 
Authors also discuss an automated compilation trajectory for complex SNN applications to their custom hybrid neuromorphic hardware.
Additionally, this is the first work that demonstrates the concept of automatic mapping of SNN models to FPGA-based neuromorphic hardware.

In \cite{adarsha_igsc}, Balaji et al. enumerate several key challenges in compiling SNNs to a neuromorphic hardware.
They show that hardware latency, especially on the shared time-multiplexed interconnect can delay some spikes more than others when reaching their destination cores.
Such delay can cause inter-spike interval (ISI) distortion and spike disorder, which lead to a significantly lower accuracy on the hardware.
Authors recommend that any SNN mapping technique needs to incorporate these metrics to ensure that the hardware accuracy is close to the accuracy that is analyzed using an application-level simulators such as Nengo~\cite{nengo}, CARLsim~\cite{carlsim}, and Brian~\cite{brian}.

%\clearpage
\noindent\textbf{Summary:} Table~\ref{tab:pbd_summary_energy} summarizes these approaches.

\begin{table}[h!]
	\renewcommand{\arraystretch}{1.2}
	\setlength{\tabcolsep}{3pt}
	\caption{Neuromorphic system software approaches targeting performance and energy optimization for platform based design.}
	\label{tab:pbd_summary_energy}
	\vspace{10pt}
	\centering
	\begin{threeparttable}
	{\fontsize{8}{10}\selectfont
	    %\vspace{-10pt}
		\begin{tabular}{u{4.0cm}|u{2.2cm}u{4.0cm}u{4.0cm}}
			\hline
			\textbf{System Software} & \textbf{Target Platform} & \textbf{Optimization Metric} & \textbf{Other Comments}\\
			\hline
			SentryOS\cite{sentryos} & $\mu$Brain\cite{mubrain} & Throughput, Utilization & Compiler and run-time manager\\
			Corelet\cite{corelet} & TrueNorth\cite{truenorth} & Core Utilization & Compiler framework \\
			LCompiler\cite{loihi_mapping} & Loihi\cite{loihi} & Core Utilization & Compiler framework \\
			PACMAN\cite{pacman}, \cite{sugiarto2017optimized} & SpiNNaker\cite{spinnaker} & Core Utilization & Compiler framework\\
			SNN-PP\cite{urgese2016optimizing}, \cite{barchi2018directed,barchi2018mapping} & SpiNNaker\cite{spinnaker} & Spike Communication Energy & Compiler and run-time manager\\
			PyNN\cite{pynn} & SpiNNaker\cite{spinnaker}, BrainScaleS\cite{brainscale}, Loihi\cite{loihi} & Core Utilization & Compiler framework \\
			Nengo\cite{nengo} & FPGA, SpiNNaker\cite{spinnaker}, Loihi\cite{loihi} & Core Utilization & Compiler framework \\
			\cite{ji2018bridge} & Tianji\cite{tianji}, PRIME\cite{prime} & Core Utilization & Compiler framework \\
			\cite{wang2021end} & Tianjic\cite{pei2019towards} & Core Utilization & Compiler framework\\
			DecomposeSNN\cite{decompose_snn}, Coreset\cite{coreset}, TraNNsformer\cite{ankit2017trannsformer} &  Crossbar-based architecture & Core Utilization & Compiler framework \\
			\cite{gao2012dynamical,neftci2013synthesizing,corradi2014mapping,nair2019mapping}, ROXANN\cite{balaji2019design} & Application-specific architecture & Dynamic Energy & Compiler framework \\
			PSOPART\cite{psopart} & Crossbar-based architecture & Spike Communication Energy & Compiler framework \\
			SpiNeMap\cite{spinemap}, PyCARL\cite{pycarl}, Neu-NoC\cite{neunoc}, \cite{balaji2019exploration,catthoor2018very,adarsha_igsc} & Crossbar-based architecture & Spike Communication Energy, Interconnect Latency, Application Accuracy & Compiler framework\\
			\cite{twisha_energy,jolpe18} & Crossbar-based architecture & Energy & Compiler framework\\
			SODASNN\cite{shihao_soda} & FPGA-based neuromorphic hardware & Energy & Compiler framework \\
			TENNLab\cite{plank2018tennlab}, \cite{mitchell2021low} & Xilinx Zynq-based Caspian\cite{mitchell2020caspian} & Energy & Compiler and run-time manager\\
			\cite{balaji2020run} & Crossbar-based architecture & Spike Communication Energy & Run-time manager\\
			\hline
	\end{tabular}}
	\end{threeparttable}
	%\vspace{12pt}
	%\vspace{-10pt}
\end{table}

\clearpage

\subsection{System Software for Thermal and Reliability Optimization}\label{sec:thermal_reliability_pbd}
We discuss key system software optimization approaches that addresses thermal and reliability aspects in executing SNN applications on a neuromorphic hardware.

In \cite{song2020case}, Song et al. propose a detailed circuit-level aging model for the CMOS transistors that are used to design the neurons in a neuromorphic crossbar with PCM synapses~\cite{shihao_igsc}. 
High voltages required to operate PCM cells can accelerate the aging in a neuron's CMOS circuitries, thereby reducing the lifetime of a neuromorphic hardware.
Authors evaluate the long-term, i.e., lifetime reliability impact of executing state-of-the-art machine learning tasks on a PCM-based neuromorphic hardware.
They show that aging is strongly dependent on temperature.
However, unlike temperature which is a short-term effect, lifetime reliability is a long-term circuit degradation and can be budgeted over time.
Therefore, maximizing the lifetime reliability via the system software avoids over-constraining a neuromorphic design and helps to reduce the reliability-related design budget.
Subsequently, authors propose to extend the SpiNeMap framework~\cite{spinemap} to incorporate neuron aging during mapping exploration.

In \cite{reneu}, Song et al. propose a framework called \underline{RENEU} to model the reliability of a crossbar in a neuromorphic hardware, considering the aging in both neuron and synapse circuitries.
Crossbar-level reliability models are then integrated to the system-level reliability model considering a Sum-of-Failure-Rates (SOFR) distribution~\cite{amari1997closed}.
RENEU incorporate three different aging mechanisms -- time-dependent dielectric breakdown (TDDB), bias temperature instability (BTI) and hot carrier injection (HCI).
This system-wide reliability model is integrated in a design-space exploration (DSE) framework involving mapping of neurons and synapses to hardware.
The key idea of RENEU is the following. 
First, it performs clustering of an SNN application using SpiNeMap~\cite{spinemap}.
Next, it uses a Hill-Climbing heuristic to map clusters to cores of the hardware incorporating the system-level reliability model. The key idea is to maximize the minimum lifetime of all crossbars in the hardware.
%Through detailed simulations authors show that by intelligently mapping neurons and synapses to the hardware, the reliability can be improved significantly.
%This leads to a longer operating lifetime of neuromorphic systems.

In \cite{gebregirogis2020approximate}, Gebregirogis et al. propose a learning and mapping approach that utilizes approximate computing for layer-wise pruning and fault-tolerant weight mapping of CNNs.
The proposed approach works as follows.
First, authors propose an approximate learning technique to remove less relevant (approximable) neurons and non-important features (weights), iteratively.
The network is trained for a few iterations to extract layer-wise error contributions of the neurons.
Subsequently, the network is retrained by pruning the approximable neurons.
Second, the pruned network is retrained to improve accuracy by fine-tuning the weights.
Finally, the retrained network is mapped to the hardware. 
To do so, a layer-wise fault tolerant memory operating voltage downscaling technique is adopted to aggressively reduce the cache supply voltage when it stores the weights of approximable layers and increase it back to the nominal value when storing the weights of non-approximable layers.
This reduces energy.
%zThe energy efficiency of the proposed approach is demonstrated on MNIST and CIFAR datasets.

In \cite{xu2021reliability}, Xu et al. propose a fault-tolerant design methodology for mapping machine learning workloads to a memristor-based neuromorphic hardware.
Authors specifically address stuck-at faults of memristive devices.
The proposed design methodology consists of the following two stages -- a general design optimization and a chip-specific design optimization.
For the general design optimization, authors propose a reliability-aware training scheme.
Here, a dropout-inspired technique and a new weighted error function are introduced to learn more robust features about stuck-at faults and their variations.
For the chip-specific design optimization, authors propose to implement the reliability-aware trained model on different ReRAM-based memristors of the hardware by exploiting the sensitivity of model weights to stuck-at faults.

In \cite{balaji2019framework}, Balaji et al. propose a framework to compute the reliability of charge pumps that are used to supply high voltages to a neuromorphic crossbar. Such high voltages are needed to operate a crossbar's PCM devices.
Authors show that if a charge pump is activated too frequently, its internal CMOS devices do not recover from stress, accelerating their aging and leading to negative bias temperature instability (NBTI) generated defects.
On the other hand, forcefully discharging a stressed charge pump can lower the aging rate of its CMOS devices, but makes the neuromorphic hardware unavailable to perform computations while its charge pump is being discharged.
The proposed framework analyzes an SNN workload using training data to identify precisely when neurons spike. Such spike information is then used to estimate the degradation in different charge pumps for a given mapping of neurons and synapses to the hardware.
The work proposes the integration of this workload-dependent aging estimation framework in a design space exploration involving distributing neurons and synapses to the hardware, thereby improving the aging in different charge pumps.

In \cite{ncrtm}, Song et al. propose \underline{NCRTM}, a run-time manager for improving the lifetime reliability of neuromorphic computing using PCM crossbars.
Due to continuous use at elevated voltages, CMOS devices in the peripheral circuit of a crossbar suffer from bias temperature instability (BTI)-induced aging~\cite{weckx2014non,das2013aging,kraak2019parametric,kraak2018degradation,das2013reliability,hebe}.
To improve reliability, it is necessary to periodically de-stress all neuron and synapse circuits in the hardware.
NCRTM is proposed to do exactly so.
It dynamically de-stresses neuron and synapse circuits in response to the short-term aging in their CMOS transistors, with the objective of meeting a reliability target.
NCRTM tracks this aging at run-time during the execution of a machine learning workload.
It de-stresses neuron and synapse circuits only when it is absolutely necessary to do so, otherwise it reduces the performance impact by scheduling de-stress operations off the critical path.

In \cite{liu2017rescuing}, Liu et al. propose a methodology to rescue bit failures in NVM-based neuromorphic hardware in order to restore the computational accuracy.
The design methodology consists of three steps.
First, authors propose to identify weights of a machine learning model that have lower impact on accuracy.
Essentially, model weights are categorized into significant and in-significant weights.
Next, authors propose a retraining algorithm to compensate for single-bit failure by re-tuning the trainable weights.
Finally, during mapping step, a redundancy mapping scheme is used to further improve the computation accuracy.

%In \cite{xu2020fault}, 

In \cite{twisha_thermal}, Titirsha et al. propose a framework to model the thermal interactions in a crossbar, which is used to map SNN models.
Using this framework, authors propose to distribute neurons and synapses to different crossbars of a hardware such that the average temperature of different crossbars is reduced, which in turn improves reliability~\cite{lifetime_reliability}.
%A crossbar's thermal formulation is based on the use of Phase Change Memory (PCM), an emerging NVM device which is used to implement the synaptic cells of the crossbar. 
The thermal formulation incorporates both the temporal component, resulting from self-heating of a PCM cell over time due to propagating spikes of a machine learning workload and the spatial component, resulting from heat transfer from nearby cells within a crossbar~\cite{das2015reliability,ukhov2012steady,das2016adaptive,ukhov2014probabilistic,das2014reinforcement,das2014temperature,das2015workload}.
The thermal formulation is integrated inside a Hill-Climbing heuristic, which is used to map neurons and synapses to different crossbars of a hardware.
Authors also show how the thermal formulation can be adapted for other NVM types.

%\cite{nittala2016toolchain}

In \cite{ahmed2021neuroscrub}, Ahmed et al. presents \underline{NeuroScrub}, a mechanism to mitigate data retention faults in NVM-based neuromorphic hardware.
Scrubbing is a technique to reprogram configuration data on to devices.
Scrubbing has been used extensively in the context of FPGA to mitigate single-event upsets (SEUs)~\cite{stoddard2016hybrid,santos2014criticality,heiner2009fpga,venkataraman2014bit,sari2013combining,das2013improving}.
NeuroScrub uses a scrubbing mechanism to counteract technology-dependent uni-directional retention faults in NVM cells used in a neuromorphic hardware.
Authors show that not all retention failures in the hardware lead to accuracy impact of inference.
Therefore, NeuroScrub only addresses uni-directional retention faults and only approximately restores intended weight matrices in the NVM memory.
To this end, authors propose to divide the weight matrix of the hidden layer into stable weights and unstable weights.
These are then mapped to separate crossbars.
Finally, scrubbing is enabled at different intervals and the accuracy impact is evaluated.

In \cite{vts_das}, Kundu et al. propose an overview of reliability issues in neuromorphic hardware and their mitigation approaches.
First, it outlines the reliability issues in a commercial systolic array-based machine learning accelerator in the presence of faults engendering from device-level non-idealities in the DRAM.
Next, it quantifies the accuracy impact of circuit-level faults in the MSB and LSB logic cones of the Multiply and Accumulate (MAC) block of the machine learning accelerator.
Finally, it presents two key reliability issues -- circuit aging and endurance in emerging neuromorphic hardware and shows the potential of SNN mapping approaches in mitigating them.

In \cite{zhang2020lifetime}, Zhang et al. propose an approach to improve the lifetime of ReRAM-based neuromorphic hardware considering thermal and aging effects.
Authors show that during programming and online training, ReRAM cells experience aging caused by high voltage operations.
Additionally, ReRAM cells with large conductance values generate large currents during programming, which change their internal and ambient temperature and thus incur thermal issues.
Thermal issues accelerate aging of ReRAM cells causing lower inference accuracy.
The proposed framework works as follows.
First, during training, aging stress on ReRAM cells is distributed relatively evenly by adjusting weights according to the current aging status of ReRAM cells.
Next, thermal effects are balanced by distributing large conductance weights across
%balancing weights corresponding to large conductance on 
the crossbar.
Finally, a row-column swapping technique is introduced during hardware mapping to deal with uneven aging and thermal effects.

In \cite{beigi2018thermal}, Beigi et al. propose a temperature-aware training and mapping of machine learning models to ReRAM-based neuromorphic hardware.
The framework operates in the following steps.
First, authors evaluate the impact of temperature on ReRAM cells of a crossbar and show how temperature variations impact accuracy.
Next, authors propose a classification approach that incorporates the temperature distribution to identify weights that have higher impact on accuracy.
Finally, a temperature-aware training process is proposed to map model weights to the hardware such that effective weights (those that have higher accuracy impact) are not mapped to hot ReRAM cells, i.e., those ReRAM cells that are more prone to generate incorrect outputs.

In \cite{zhang2019aging}, Zhang et al. propose an algorithm-software co-optimization for mapping vector-matrix computations in deep neural networks to mitigate limited endurance of memristors in a neuromorphic hardware.
Authors show that during execution of deep learning models on the hardware, memristors need to be repeatedly tuned, i.e., reprogrammed using a pulse of very high voltage.
This high voltage may cause a change in the filament inside a memristor, leading to a degradation of the valid range of its conductance and thus the number of usable conductance levels.
If trained weights are mapped to the hardware assuming a fresh state of the memristor, the target conductance might fall outside of the valid range and therefore, the programmed conductance may deviate from the target conductance.
The proposed mapping framework consists of the following two steps.
First, a skewed-weight training is used for deep learning models to deal with such reliability issues.
Essentially, the idea is to train deep learning models using smaller weights only.
This is because
smaller weights require smaller conductance and equivalently, larger resistance on the memristor cell.
This reduces the amount of current flowing through the cell, which improves its reliability.
Finally, when mapping weights of a deep learning model to memristors, the current status of the memristors are taken into account. This status is tracked using a representative memristor.
During mapping, if memristors are stressed, then they are programmed with smaller conductance values.
Otherwise, the original conductance value is programmed.

In \cite{twisha_endurance}, Titirsha et al. propose a tradeoff analysis involved in mapping of SNNs on a crossbar-based neuromorphic hardware where NVM cells are used for synaptic storage.
Authors show that a major source of voltage drop in a crossbar are the parasitic components on bitlines and wordlines, which are deliberately made longer to achieve lower cost-per-bit.
When mapping SNNs to a crossbar, cells on shorter current paths are faster to access (high performance) but have lower write endurance due to high currents.
On the other hand, cells on longer current paths are slower to access but have higher endurance due to small currents.
This reliability and performance formulation is incorporated in an SNN mapping framework.
The framework first partitions an SNN into clusters using SpiNeMap~\cite{spinemap}.
Subsequently, clusters are mapped to crossbars by exploiting the endurance-performance tradeoff.
The overall objective is to balance the endurance of NVM cells in a crossbar with the minimum performance degradation.

In \cite{espine}, Titirsha et al. propose \underline{eSpine}, an approach to mitigate the limited write endurance of NVM cells of a neuromorphic crossbar using intelligent mapping of neurons and synapses to the hardware.
Authors show that due to large parasitic components on bitlines and wordlines, there is a significant variation of the current propagating through different NVM cells in a crossbar.
Through a detailed circuit-level modeling using PCM cells, authors show the endurance variation in a crossbar generated due to the current variation.
eSpine operates in two steps.
First, it partitions an SNN into clusters of neurons and synapses using the Kernighan-Lin Graph Partitioning algorithm ensuring that a cluster can be mapped to a crossbar of the hardware.
Next, it uses PSO to map clusters to cores.
Across different PSO iterations, eSpine maximizes the minimum write endurance of different crossbars.
Within each PSO iteration, eSpine intelligently programs synaptic weights to different PCM cells of a crossbar such that
those synapses that are activated too frequently (i.e., those that have more spikes) within a workload are mapped to PCM cells that have high write endurance, and vice verse.
%intelligently placing synapses of a cluster to PCM cells of a crossbar in each PSO iteration such that those synapses that are activated too frequently within a workload are mapped on PCM cells that have high write endurance. 
In this way, eSpine balances the endurance within each crossbar of a neuromorphic hardware, simultaneously maximizing the overall write endurance.

In \cite{song2021improving}, Song et al. propose an approach to mitigate read disturb issues of ReRAM cells in a neuromorphic hardware using intelligent synapse mapping strategies.
Authors show that an ReRAM cell can switch its state after reading its content a certain number of times. 
Such behavior challenges the integrity and program-once-read-many-times philosophy of implementing machine learning inference on neuromorphic systems, impacting the Quality-of-Service (QoS).
To address read disturb issues, authors first characterize read disturb-related endurance of ReRAM cells in a neuromorphic hardware.
They categorize ReRAM cells into strong and weak cells, where the strength of an ReRAM cell is measured as a function of its read endurance; higher the read endurance, stronger is the cell.
Authors then propose  an intelligent synapse mapping technique, which analyzes spikes propagating through each synapse of a machine learning model during inference and uses such information to map the model's synaptic weights to the ReRAM cells of a crossbars considering the variation in their read endurance values.
The overall objective is the following: those synapses that propagate more spikes are mapped on stronger ReRAM cells, i.e., cells that can sustain more read operations.
In this way, weaker cells are not overly stressed, which improves lifetime.

In \cite{pauldt}, Paul et al. propose an extension of the read disturb mitigation approach of ~\cite{song2021improving}.
Fist, the proposed framework models the performance overhead in periodically reprogramming model parameters to the ReRAM cells of a neuromorphic hardware in order to mitigate their read disturb issues.
Second, it exploits machine learning model characteristics to identify non-critical model parameters, i.e., those that have no impact on accuracy and eliminate them from the critical path of deciding the reprogramming interval.
Third, it uses a convex optimization formulation of cluster mapping to crossbar in order to reduce the system overhead.

\clearpage

\noindent\textbf{Summary:} Table~\ref{tab:pbd_summary_reliability} summarizes these approaches.

\begin{table}[h!]
	\renewcommand{\arraystretch}{2.0}
	\setlength{\tabcolsep}{3pt}
	\caption{Neuromorphic system software approaches targeting thermal and reliability optimization for platform based design.}
	\label{tab:pbd_summary_reliability}
	\vspace{10pt}
	\centering
	\begin{threeparttable}
	{\fontsize{8}{10}\selectfont
	    %\vspace{-10pt}
		\begin{tabular}{u{4.0cm}|u{2.2cm}u{4.0cm}u{4.0cm}}
			\hline
			\textbf{System Software} & \textbf{NVM Technology} & \textbf{Reliability Issues} & \textbf{Other Comments}\\
			\hline
			\cite{song2020case,shihao_igsc,balaji2019framework,vts_das}, RENEU\cite{reneu}, NCRTM\cite{ncrtm} & PCM & Aging in CMOS transistors of neuron circuitry & Time-Dependant Dielectric Breakdown (TDDB), Bias Temperature Instability (BTI)\\
			\cite{twisha_thermal} & PCM & Self-heating  temperature of PCM cells & Minimize the average temperature of a neuromorphic hardware \\
			\cite{twisha_endurance}, eSpine\cite{espine} & PCM & Limited write endurance of PCM cells & Incorporate bitline and wordline parasitics\\
			\cite{xu2021reliability,liu2017rescuing,zhang2020lifetime,zhang2019aging}, NeuroScrub\cite{ahmed2021neuroscrub} & ReRAM & Soft and hard NVM manufacturing issues & --\\
			\cite{beigi2018thermal} & ReRAM & Temperature in a crossbar & -- \\
			\cite{song2021improving,pauldt} & ReRAM & Read disturb issues in ReRAM cells in a neuromorphic crossbar & -- \\
			\cite{gebregirogis2020approximate} & -- & Voltage downscaling impact on SRAM-based synaptic memory & -- \\
			\hline
	\end{tabular}}
	\end{threeparttable}
	%\vspace{12pt}
	%\vspace{-10pt}
\end{table}

%\clearpage

\subsection{Application and Hardware Modeling for Predictable Performance Analysis}\label{sec:pbd_abstract}
The design of a neuromorphic platform is getting more and more complex. 
To manage the design complexity, a predictable design flow is needed. 
The result should be a system that guarantees that an SNN application can perform its operations within the strict timing deadlines. 
This requires that the timing behavior of the hardware, the software, as well as their interaction can be predicted.
We discuss these design flows.

In \cite{sdfsnn}, Das et al. propose an approach called \underline{SDFSNN}, which uses Synchronous Dataflow Graphs (SDFGs)~\cite{lee1987synchronous} for mapping SNNs to a neuromorphic hardware.
SDFGs are commonly used to model streaming applications that are implemented on a multi-core system~\cite{SB00,jiashu2012design,das2018reliable}. 
Both pipelined streaming and cyclic dependencies between tasks can be easily modeled in SDFGs.
These graphs are used to analyze a system in terms of key performance properties such as throughput, execution time, communication bandwidth, and buffer requirement~\cite{Stuijk06dac,singh2013rapiditas,das2012faultRSP}.
Authors show that SDFSNN can be used to model both feed-forward and recurrent neural networks.
Using SDFG semantics, authors model an SNN as an application graph and a many-core neuromorphic hardware as a platform graph.
Using the mapping framework of SDFG~\cite{stuijk2007predictable,das2018reliable}, authors propose to distribute neurons to cores and estimate the corresponding throughput.
%show that SDFSNN can be used to distribute neurons and synapses to the hardware, balancing the workload on different cores. By binding the SDFG to a neuromorphic hardware, SDFSNN allows estimating throughput of the hardware implementation of different machine learning models.

In \cite{sdfsnn_pp}, Balaji et al. automated the generation of SDFG from an SNN using a framework called \underline{SDFSNN++}. 
Nodes of an SDFG are called {actors}, which
are computed by reading {tokens} from their input ports and writing the results of the computation as tokens on output ports.
Port rates are visualized as annotations on edges. Actor execution is also called {firing}, and it requires a fixed amount of time to execute. Edges in the graph are called {channels} and they represent dependencies among actors.
An actor is said to be {ready} when it has sufficient input tokens on all its input channels and sufficient buffer space on all its output channels; an actor can only fire when it is ready.
Table~\ref{tab:snn_sdfg_mapping} shows the one-to-one mapping of an SNN to SDFG properties. 
SDFSNN++ uses a real-time calculus~\cite{thiele2000real} to estimate the throughput of an SDFG representation of an SNN mapped to a platform with unbounded resources.

\begin{table}[h!]
	\renewcommand{\arraystretch}{2.0}
	\setlength{\tabcolsep}{3pt}
	\caption{One -to-one mapping of SNN to SDFG terminology.}
	\label{tab:snn_sdfg_mapping}
	\vspace{10pt}
	\centering
	\begin{threeparttable}
	{\fontsize{8}{10}\selectfont
	    %\vspace{-10pt}
		\begin{tabular}{c|c}
			\hline
			\textbf{SDFG Terminology} & \textbf{SNN Terminology}\\
			\hline
			actor & neuron\\
			channel & synapse \\
			token & spike \\
			\hline
	\end{tabular}}
	\end{threeparttable}
	%\vspace{12pt}
	%\vspace{-10pt}
\end{table}

In \cite{dfsynthesizer}, Song et al. propose a framework called \underline{DFSynthesizer}, which is used to map SNN models to state-of-the-art neuromorphic hardware with a limited number of cores.
With such limitations, hardware resources such as the processing cores may need to be shared (time-multiplexed) across the neurons and synapses of an SNN model, especially when the model is large and cannot fit on a single core of the hardware.
Time-multiplexing of hardware resources introduces delay, which may lead to violation of real-time requirements.
To address this, DFSynthesizer first partitions an SNN model into clusters, where each cluster can fit onto a core of the hardware. 
Next, clusters are mapped to cores, where each core may need to execute multiple clusters.
To do this mapping, authors propose to use Max-Plus Algebra~\cite{akian2006max}, which determines the best way clusters need to be distributed in order to maximize the throughput.
Finally, a schedule is constructed based on Self-Timed Execution~\cite{DasKV14} to sequence the execution of clusters on each core.

In \cite{dfsynthesizer_pp}, Song et al. extended the DFSynthesizer framework of~\cite{dfsynthesizer}. 
The proposed framework called \underline{DFSynthesizer++} is an end-to-end framework for synthesizing machine learning programs to the hardware, improving both throughput and energy consumption.
The framework first generates a machine learning workload by analyzing the training data.
It then partitions the workload into clusters using a greedy graph partitioning algorithm. 
Next, it distributes clusters to the hardware, time-multiplexing hardware resources to maximize throughput.
Finally, it sequences clusters that are mapped to the same core such that the generated sequence guarantees performance.
DFSynthesizer++ introduces the following functionalities to allow for the conversion of CNN architectures such as LeNet, AlexNet, and VGGNet to their spiking counterparts.
\begin{enumerate}
    \item \emph{1-D Convolution:} The 1-D convolution is implemented to extract patterns from inputs in a single spatial dimension. A 1xn filter, called a kernel, slides over the input while computing the element-wise dot-product between the input and the kernel at each step.
    \item \emph{Residual Connections:} Residual connections are implemented to convert the residual block used in CNN models such as ResNet. Typically, the residual connection connects the input of the residual block directly to the output neurons of the block, with a synaptic weight of `1'. This allows for the input to be directly propagated to the output of the residual block while skipping the operations performed within the block.
    \item \emph{Flattening:} The flatten operation converts the 2-D output of the final pooling operation into a 1-D array. This allows for the output of the pooling operation to be fed as individual features into the decision-making fully connected layers of the CNN model.
    \item \emph{Concatenation:} The concatenation operation, also known as a merging operation, is used as a channel-wise integration of the features extracted from 2 or more layers into a single output.
\end{enumerate} 

In \cite{shihao_designflow}, Song et al. propose a complete design flow (roughly based on the design flow proposed for embedded multiprocessor systems~\cite{clement1999fast,jiashu2012design,stuijk2010predictable}.) for mapping throughput-constrained SNN applications to a neuromorphic hardware.
The design flow explores the tradeoff between buffer size on each core and throughput.
The design flow wors as follows.
First, it uses Kernighan–Lin graph partitioning heuristic to create SNN clusters such that each cluster can be mapped to a core of the hardware.
The partitioning approach minimizes inter-cluster spike communication, which improves latency and reduces communication energy.
Next, it maps clusters to cores using an instance of the PSO algorithm, while exploring the design space of throughput and buffer size.
Specifically, for each unit of buffer size on hardware cores, it finds the best throughput possible with the given buffer allocation using the DFSynthesizer-based mapping framework~\cite{dfsynthesizer_pp}.
The algorithm is iterated for every increment of buffer size on cores. 
In this way, the entire design space of throughput and buffer size requirement is explored.
The proposed design flow allows system developers to select a buffer size configurations
required to achieve a given throughput performance.

%\subsection{Summary}

\section{Hardware-Software Co-Design}\label{sec:hsco}
Most electronics systems consists of a hardware platform which executes software programs.
Hardware-software co-design is a system design paradigm where system-level objectives such as cost, performance, power, and reliability are met by exploiting the synergism of hardware and software through their concurrent design and optimization~\cite{wolf1994hardware,de1997hardware,staunstrup1997hardware,de2002readings}.
Similar to many electronics system design~\cite{balarin1997hardware,das2013aging,li2000hardware,bolsens1997hardware,das2015hardware}, hardware-software co-design is also used for the design of neuromorphic systems~\cite{rajendran2019building}.
Here, we review the software technology for hardware-software co-design.\footnote{Hardware-software co-design methodologies are also proposed for CNN accelerators~\cite{chae2020centralized,lee2020architecture,kim2020hardware,wu2021software,liu2018memristor,li2021real,cheng2021future,sunny2021crosslight,hanif2022cross,bringmann2021automated,li2018network,kwon2018co,zhu2019sparse,baharani2021deepdive,amid2019co,hao2018deep,guo2016model,hao2021enabling,zhou2021analognets,jayakodi2021general,ambrosi2018hardware}.
To keep the focus on neuromorphic computing, we only review methodologies for spiking-based neuromorphic systems.}

In \cite{neutrams}, Ji et al. propose \underline{NEUTRAM} (Neural Network Transformation, Mapping, and Simulation), a co-design methodology for two target neuromorphic hardware platforms -- Tianji~\cite{tianji} and PRIME~\cite{prime}.
The design methodology works in three steps.
In step 1, applications are represented using a high-level language such as PyNN~\cite{pynn} and CARLsim~\cite{carlsim}.
In step 2, the SNN model is transformed to incorporate hardware constraints.
The transformation algorithm divides an existing SNN into a set of simple network units and retrains each unit iteratively.
In step 3, the transformed SNN is mapped to the hardware using a runtime tool.
Using this co-design methodology, authors show the tradeoffs between network error rates and hardware consumption.
%This includes incorporating hardware constraints during backpropagation, 

In \cite{zhao2019hardware}, Zhao et al. propose a co-design methodology for designing a neuromorphic hardware for optical neural networks (ONNs)~\cite{sui2020review,lu1989two,shariv1989all,stoll1988continuous,wang2022optical}.
Conventionally, an optical hardware is synthesized from a software-trained network.
The proposed methodology adopts a different route: the optical hardware implementation and the software training implementation are co-designed to reduce the hardware implementation cost.
Authors first propose an area-efficient architecture of ONN, which includes a sparse tree network block, a single unitary block and a diagonal block for each neural network layer. 
Next, authors propose the software embodiment of these hardware structures.
Finally, the hardware and software are co-optimized to reduce the area cost.

In \cite{neuroxplorer}, Balaji et al. propose \underline{NeuroXplorer}, a hardware-software co-design methodology for implementing SNNs on a neuromorphic hardware. %incorporating SNN-based applications and algorithms, a system software to map SNN applications to the hardware, and a many-core neuromorphic hardware integrating NVMs.
The key idea of NeuroXplorer is to optimize the system software, e.g., the compilation framework, cluster generation, and CNN conversion alongside the hardware architecture.
Authors show that NeuroXplorer can design the hardware, including the number of cores, number of neurons per core, synaptic capacity of each core, interconnect configuration, and routing algorithm for a given SNN application.
NeuroXplorer can also optimize the architecture of each core, such as generating two-layer crossbar architecture, three-layered $\mu$Brain architecture, and the decoupled architecture consisting of separate synaptic memory.
For each of these design choices, NeuroXplorer can simultaneously optimize the cluster generation and the mapping technology.

In~\cite{paul_mdpi_2022}, Paul et al. propose a co-design methodology to implement respiratory anomaly detection of newborn infants on neuromorphic systems.
The methodology works as follows. 
First, it optimizes the CNN architecture to optimize the classification accuracy.
Next, it converts the CNN into spiking architecture using the conversion framework proposed in~\cite{jolpe18}. 
Finally, it performs optimization of the system software and hardware using the NeuroXplorer framework~\cite{neuroxplorer}.

In \cite{plank2017unified}, Plank et al. propose a hardware-software co-design framework similar to NeuroXplorer~\cite{neuroxplorer}.
In addition to architectural explorations, the proposed framework can also co-design the NVM cells that are used for synaptic storage.
The design framework is unified and therefore, new results on device can be applied to application instantly and estimate the performance impact.
%This allows the co-design of application and hardware.

%\cite{yantir2022hardware}

In \cite{an2021three}, An et al. propose a hardware-software co-design methodology to design three-dimensional (3D) neuromorphic hardware with two layers of memristive synapses.
The hardware-software co-design methodology combines SNN training algorithm, Whetstone~\cite{whetstone} with 3D integrated circuits (3D-ICs)~\cite{shulaker2015monolithic,tu2011reliability,lee2012architecture}.
The co-design is performed in three steps.
First, two neural networks (MLP and CNN) are trained using Whetstone.
Next, during the sharpening procedure, weights of the SNN are stored and mapped to memristors (ReRAM) in the binary format.
Finally, memristors are fine-tuned to the performance of the neuromorphic system using the NeuroSim simulator~\cite{neurosim}.
The co-design framework optimizes for 1) the training accuracy by fine-tuning the weights of an SNN and 2) heat diffusion effect in memristors by adding heat dissipation layers. 
The objective is to co-optimize design area, power, and latency.

In \cite{shi2020learned}, Shi et al. propose a co-design approach to design specialized hardware and software for deep learning accelerators.
The essential idea of the framework is a design space exploration (DSE) framework that involves two exploration loops.
The outer loop optimizes the hardware architecture, while the inner loop optimizes the software mapping to the architecture.
These explorations are performed using a nested Bayesian Optimizer that uses Bayesian models of hardware and software performance to guide the search process.
This is similar to the design space exploration frameworks proposed for multi-core/multiprocessor designs~\cite{pimentel2016exploring,singh2013rapiditas,lahiri2004design,piscitelli2012design,das2014combined,lee2020hardware}.
Through this DSE, authors show a significant opportunity to improve the energy-delay product.
Although targeted for ANN accelerators, the methodology can be applied also for SNN hardware.

In \cite{schuman2020evolutionary}, Schuman et al. propose \underline{EONS} (Evolutionary Optimization for Neuromorphic Systems) for rapid prototyping of SNN applications on neuromorphic systems~\cite{schuman2016evolutionary}.
EONS operates in four steps -- generation, evaluation, selection, and reproduction.
The initial population is randomly generated.
Next, the network is evaluated in population using a fitness score.
Next, a network is selected as parent using an evolutionary algorithm.
Finally, reproduction operations such as duplication, crossover, and mutation are performed on the parent network.
EONS can be applied to any tasks such as classification and control, without changing the underlying algorithm.
EONS can be used to map an SNN application to an existing neuromorphic platform such as Caspian~\cite{mitchell2020caspian} by incorporating its hardware constraints, essentially a platform-based design concept.
EONS can also perform optimization of the hardware, e.g., a reservoir in a liquid state machine for a given application.
Therefore, EONS can co-design both the hardware and software, exploring the design tradeoffs.

%\cite{chae2020centralized}

%\textcolor{red}{
In \cite{li2021multi}, Li et al. propose a co-design methodology to design an accurate and communication-optimized Liquid State Machine (LSM) architecture.
LSM works on the principle of reservoir computing~\cite{lukovsevivcius2009reservoir}. 
An LSM consists of a reservoir of recurrently-connected spiking neurons and a supervised readout~\cite{maass2011liquid,HeartEstmNN}.
The co-design methodology consists of three engines.
The first is the multi-objective LSM architecture search engine involving the exploration of the number of neurons in the liquid and their connectivity.
The second is the NoC architecture search engine involving the number of nodes, NoC topology and routing.
The third is the mapping engine, which maps the LSM to the NoC nodes.
The objective of the co-design is to efficiently implement LSM architectures to a NoC-based neuromorphic hardware.
%}

In \cite{gopalakrishnan2020hfnet}, Gopalakrishnan et al. propose a co-design approach for CNN architecture implemented on crossbar arrays by exploring the design space of different hardware-friendly convolution techniques and their software-based mapping approaches.
The target neuromorphic hardware consists of crossbars that are interconnected using a shared interconnect.
The iterative co-design methodology works in the following steps.
First it initiates the number of convolutional layers and the number of depthwise separable convolution layers.
Next, it decides on the number of feature maps per layer, with
the constraint of avoiding core matrix splitting.
To do so, it puts a constraint on the fan-in degree of a neuron.
Finally, the optimized CNN (called HFNet) is mapped to obtain the minimum number of cores and estimate the accuracy.
The process is repeated to explore the design space of accuracy and hardware area.

%In \cite{yantir2022hardware,lee2020architecture}, Lee et al. propose accuracy-aware design optimization of neural networks implemented on ReRAM-based neuromorphic hardware.

%\cite{kim2020hardware}

%\cite{wu2021software}

%\cite{sengupta2021architecture}

%\cite{lee2020hardware}

%\cite{liu2018memristor}

In \cite{fang2021neuromorphic}, Fang et al. propose neuromorphic learning and hardware co-design for temporal pattern learning.
Authors first propose an efficient training algorithm for SNN with LIF neurons to learn from temporal data.
The learning algorithm allows to process temporal data without using a recurrent structure, which reduces the design complexity.
Several design strategies are proposed to reduce the design complexity while achieving comparable accuracy.
Next, the learning algorithm is co-designed on a ReRAM-based neuromorphic hardware.

%\cite{li2021real}

In \cite{schuman2021software}, Schuman et al. propose a software framework to evaluate the impact of four SNN learning algorithms on three neuromorphic simulators and demonstrate the same using four simple classification tasks.
The training algorithms evaluated for SNNs include Decision Tree, EONS~\cite{schuman2016evolutionary}, Whetstone~\cite{whetstone}, and Reservoir Computing~\cite{lukovsevivcius2009reservoir}.
The neuromorphic platforms evaluated are Caspian~\cite{mitchell2020caspian}, Generic Neuromorphic Processor (GNP), and NEST~\cite{nest}.
Authors evaluate the hardware area and training accuracy through these explorations.
Authors demonstrate the importance of learning algorithm and hardware co-design for neuromorphic computing.

%\cite{hao2021enabling}

%\cite{zhou2021analognets}

%\cite{jayakodi2021general}

In \cite{oltra2021hardware}, Oltra-Oltra et al. propose a hardware-software co-design for distributed SNN architectures.
First, authors propose a Single Instruction Multiple Data (SIMD)-based SNN architecture.
Next, they develop a tool chain for mapping SNN models to the target architecture.
Through a co-design framework called \underline{HEENS} (Hardware Emulator of Evolvable Neural Spiking), authors propose to jointly optimize the architecture and its tool chain to achieve maximum benefit from the SIMD design and their use for distributed neuromorphic computing.

In \cite{frenkel2021bottom}, Frenkel et al. propose a co-design methodology for hardware implementation of SNNs.
First, authors propose analog, mixed-signal and digital circuit design styles for neuromorphic hardware.
They exploit the boundary between processing and memory through the integration of time multiplexing, in-memory computation, and novel devices.
Next, they evaluate tradeoffs for two alternate implementation styles -- a bottom-up design style, where the hardware is designed first and then used to map an SNN application, and a top-down style, where an SNN application is first optimized and then implemented in hardware.
The proposed co-design methodology involves integrating the top-down and bottom-up design styles to jointly optimize both application and hardware.

%\cite{cheng2021future}

%\cite{sunny2021crosslight}

%\cite{hanif2022cross}

%\cite{bringmann2021automated}

%\cite{li2018network}

%\cite{kwon2018co}

%\cite{zhu2019sparse}

%\cite{baharani2021deepdive}

%\cite{amid2019co}

%\cite{hao2018deep}

%\cite{guo2016model}

%\cite{johnston2010fpga}

%\cite{ambrosi2018hardware}

%\cite{zhang2020mamap}

In \cite{ankit2017resparc}, Ankit et al. propose RESPARC, a reconfigurable platform for neuromorphic hardware using memristive crossbars.
RESPARC is a three-tiered reconfigurable architecture.
Tier 1 consists of the \textit{Macro Processing Engine}, a reconfigurable compute unit to map neurons with variable fanins.
Tier 2 consists of \textit{NeuroCell}, a reconfigurable datapath to map SNNs with varying inter- and intra-layer connections.
Tier 3 consists of the RESPARC, the reconfigurable neuromorphic core to map SNNs with varying sizes.
Through co-design of these layers, authors show the design space exploration opportunities and the significant energy improvements possible in implementing SNNs on a neuromorphic hardware.

%\section{Design-Technology Co-Optimization}\label{sec:dtco}
%Design-technology co-optimization (DTCO) and more generally, system-technology co-optimization is a system design paradigm to enable system-level objectives through co-optimization of application, software, hardware, and technology. 
%DTCO is a key component to enable technology scaling for advanced process technology nodes.
%As for other system designs, for neuromorphic computing technological choices are complex.

In \cite{mallik2017design}, Mallik et al. propose a design-technology tradeoff analysis to implement SNNs on ReRAM cells of a neuromorphic hardware.
Authors show that the requirement to have multiple distinct levels is the key bottleneck for using ReRAM as synaptic cells in a neuromorphic hardware.
Accordingly, authors propose a mixed-radix encoding for multi-level ReRAM cells.
Authors show the tradeoff between single-level and multi-level cells in terms of their design and technology using silicon data and also evaluate their impact on classification accuracy.
Subsequently a design-technology co-optimization is needed to implement SNN applications on ReRAM-based neuromorphic systems.

In \cite{ankita_igsc}, Paul et al. propose a design-technology co-optimization in implementing SNNs on NVM-based neuromorphic hardware.
Using circuit-level simulations with ReRAM technology, authors show the negative impact of technology scaling on the read endurance of ReRAM cells.
Specifically, they show that at scaled technology nodes, the value of parasitic components on bitlines and wordlines of a crossbar increases, which create a significant variation in the read endurance of ReRAM cells of a crossbar.
However, technology scaling offers benefits such as improvement of integration density and cost-per-bit.
Authors propose a design flow that incorporates technological characteristics in their software mapping flow so that neurons and synapses can be placed in a crossbar mitigating the negative impact of technology scaling.
%The key idea is to map synapses that communicate more spikes onto NVM cells that have higher read endurance for a given technology node.

%\cite{yan2017understanding}

%\cite{ankit2018neuromorphic}

In \cite{ku2018design}, Ku et al. propose a design-architecture co-optimization to improve the area efficiency of LSM-based speech recognition with monolithic 3D technology.
The exploration involves 1) ASIC implementation of LSM in 2D and 3D-IC technologies and their are area comparison, 2) comparison of synapse models and memory distribution on power, performance, and area.
The design is implemented in 28nm node.
For 3D-IC design, the co-optimization framework uses a hierarchical Shrunk-2D design flow.
In this design flow, a pseudo 3D design is built from 2D, by scaling down cell dimension, wire pitch, and wire width by a factor of \ineq{\sqrt{2}}.
Actual implementation-wise, individual neuron and top-level cells are partitioned into two tiers.
First, the design methodology starts with a top-level floorplanning, which is based on the shrunk geometry and timing budget.
Then for each neuron, it builds a two-tier folded design using the Shrunk-2D flow.
Finally, the tier-by-tier routing of the design is performed and netlist generated.

%\cite{cai2020technology}

%\cite{niu2019design}

In \cite{chen2015technology}, Chen et al. propose a technology-design co-optimization for resistive crosspoint array-based neuromorphic hardware.
Several key design and technological enhancements are proposed including larger wire width to reduce IR drop, the use of multiple cells for each synaptic element to mitigate device-to-device variations, and fully-parallel read and write schemes of the weighted sum and weight updates in the crosspoint array.
Authors evaluate area, energy, and latency tradeoffs for these enhancements involving array design and resistive technology (ReRAM) at 65nm technology node.

%\cite{liu2021system}

%\cite{li2015merging}

%\cite{nguyen2020fully}

%In \cite{rajendran2019building}, Rajendran et al. enumerate the challenges in co-optimization of algorithm, architecture and nanoscale memristive devices for neuromorphic computing.

%\cite{takeuchi2021memory}

%\cite{wang2021multi}

%\cite{amrouch2021cross}

%\cite{cao2021study}

%\cite{wang2021optimizing}

\section{Outlook}\label{sec:outlook}
Over the past decade, neuromorphic computing has seen significant progress on the silicon technology, hardware, and software fronts. 
This is primarily due to the questionable future of Moore's law and the growing demand for brain-like functionality such as visual and auditory scene analysis and reasoning.
However, today's neuromorphic hardware nodes can perform several different types of scientific computations and not just limited to machine learning. 
It is unclear how such scientific computations can be efficiently mapped to event-driven operations supported on a neuromorphic hardware using existing software technologies.
This is particularly important if neuromorphic computing is to be integrated into the existing computing workflows involving CPUs and GPUs. 
%Furthermore, the software technology also needs to be  if neuromorphic 

In the future, neuromorphic systems are expected to aggregate multiple heterogeneous neuromorphic hardware nodes to generate a massively parallel system that can solve scientific computations that are far too complex for a single-node neuromorphic hardware.
However, despite the significant technological advances made on the software front, it remained to be seen how portable are existing software technologies to such large-scale systems.
Finally, virtualization of neuromorphic systems opens up a new avenue for research on software technologies.

\section*{Acknowledgement}
This material is based upon work supported by the U.S. Department of Energy under Award Number DE-SC0022014 and by the National Science Foundation under Grant
Nos. CCF-1942697 and CCF-1937419.

\section*{References}
\bibliographystyle{IEEEtran}
%\IEEEtriggeratref{30}
\bibliography{disco,external}

\end{document}